\documentclass[10pt]{article}

\usepackage[authoryear]{natbib}
\usepackage{graphicx}
\usepackage{subcaption}
\usepackage{booktabs}
\usepackage{multirow}
\usepackage{amsmath}
\usepackage{bm}
\usepackage{amsthm}
\usepackage{amsfonts}
\usepackage{amssymb}
\usepackage{mathtools}
\usepackage{algorithmic}
\usepackage{float}
\usepackage[
    vlined,
    ruled,
    linesnumbered
]{algorithm2e}
\usepackage{xcolor}
\usepackage[hidelinks]{hyperref}
\usepackage{cleveref}
\usepackage{takeuchi_mcr}
\usepackage{lipsum}

\newtheorem{theorem}{Theorem}[section]

\title{\vspace{0mm}Statistical Testing Framework \\for Clustering Pipelines by Selective Inference}
\date{\today}

\makeatletter
\def\@fnsymbol#1{\ensuremath{\ifcase#1\or
{1}\or
{2}\or
{\ast}\or
{\dagger}\or
\else\@ctrerr\fi}}
\makeatother

\author{
Yugo Miyata\thanks{Nagoya University}
\and
Shiraishi Tomohiro\footnotemark[1] \thanks{RIKEN}
\and
Nishino Shuichi\footnotemark[1] \footnotemark[2]
\and
Ichiro Takeuchi\footnotemark[1] \footnotemark[2] \thanks{Corresponding author. e-mail: takeuchi.ichiro.n6@f.mail.nagoya-u.ac.jp}
}

\begin{document}

\maketitle
\thispagestyle{empty}

\begin{abstract}
    \noindent
    A data analysis pipeline is a structured sequence of steps that transforms raw data into meaningful insights by integrating multiple analysis algorithms.
In many practical applications, analytical findings are obtained only after data pass through several data-dependent procedures within such pipelines.
In this study, we address the problem of quantifying the statistical reliability of results produced by data analysis pipelines.
As a proof of concept, we focus on clustering pipelines that identify cluster structures from complex and heterogeneous data through procedures such as outlier detection, feature selection, and clustering.
We propose a novel statistical testing framework to assess the significance of clustering results obtained through these pipelines.
Our framework, based on selective inference, enables the systematic construction of valid statistical tests for clustering pipelines composed of predefined components.
We prove that the proposed test controls the type I error rate at any nominal level and demonstrate its validity and effectiveness through experiments on synthetic and real datasets.

\end{abstract}

\newpage
\section{Introduction}
\label{sec:introduction}

In practical data-driven decision-making tasks, integrating various types of data analysis steps is crucial for addressing diverse challenges.
For instance, in genetic research aimed at identifying subgroups of patients with a specific disease, the process often begins with preprocessing tasks such as outlier removal and dimensionality reduction.
This is followed by applying clustering algorithms to identify patient subgroups, and then testing whether there are significant differences in clinical outcomes between the identified groups.
Such a systematic and structured sequence of data processing and analysis steps is referred to as a \emph{data analysis pipeline}, which plays a pivotal role in ensuring the reproducibility and reliability of data-driven decision-making.

\begin{figure}[htbp]
    \centering
    \includegraphics[width=0.8\linewidth]{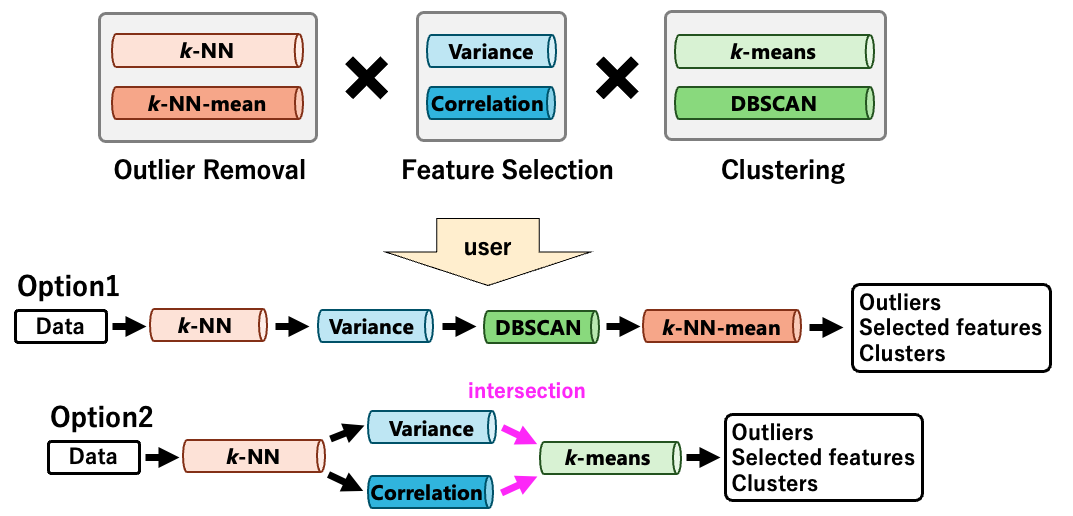}
    \caption{
    Two examples of clustering pipelines composed of outlier detection (OD),
    feature selection (FS), and clustering components,
    for which the proposed framework provides statistically valid $p$-values
    without additional implementation effort.
    }
    \label{fig:two_examples}
\end{figure}

In this study, as an example of data analysis pipelines, we consider a class of \emph{clustering pipelines} that integrates various outlier detection (OD) algorithms, feature selection (FS) algorithms, and clustering algorithms.
Figure~\ref{fig:two_examples} shows examples of two such pipelines.
The pipeline on the top (\texttt{option1}) starts with $k$-NN ($k$-nearest-neighbor)-based outlier removal, followed by variance-based feature selection, and then applies DBSCAN (Density-Based Spatial Clustering of Applications with Noise) clustering algorithm to obtain the initial cluster assignments; it concludes with a second round of $k$-NN-mean ($k$-nearest-neighbor-mean)-based outlier removal within each cluster.
The pipeline on the bottom (\texttt{option2}) begins with $k$-NN based outlier removal, continues with feature selection based on both correlation and variance criteria, and applies $k$-means clustering algorithm to obtain the final cluster assignments.

When a data-driven approach is used for high-stakes decision-making tasks such as medical diagnosis and personalized treatment, it is crucial to quantify the reliability of the final results by considering all steps in the pipeline.
The goal of this study is to develop a statistical test for a specific class of clustering pipelines, allowing the statistical significance of cluster-based findings obtained through the pipeline to be properly quantified in the form of $p$-values.
The first technical challenge in achieving this is the need to appropriately account for the complex interrelations between pipeline components to determine the overall statistical significance.
The second challenge is to develop a universal framework capable of performing statistical tests on arbitrary pipelines (within a given class) rather than creating individual tests for each pipeline.

To address these challenges, we introduce the concept of selective inference (SI)~\citep{taylor2015statistical,fithian2015selective,lee2014exact}, a novel statistical inference approach that has gained significant attention over the past decade.
The core idea of SI is to characterize the process of selecting hypotheses from the data and calculate the corresponding $p$-values using the sampling distribution, conditional on this selection process.
We propose an approach based on SI that provides valid $p$-values for any clustering pipeline configuration within the aforementioned class.
We also introduce a software implementation framework that supports SI for any pipeline configuration within this class without requiring additional implementation efforts.
Specifically, with our framework, the statistical significance of cluster-based findings from any pipeline in this class can be quantified as valid $p$-values, with no extra implementation required beyond specifying the pipeline.

We note that our long-term goal beyond this current study is to ensure the reproducibility of data-driven decision-making by accounting for the entire pipeline from raw data to the final results, with the current study on a class of clustering pipelines serving as a proof of concept for that goal.

\clearpage

\paragraph{Related Work.}

Most research on data analysis pipelines is concentrated in the field of software engineering rather than machine learning~\citep{sugimura2018building,hapke2020building,drori2021alphad3m}, with a primary focus on the design, implementation, testing, and maintenance of pipeline systems to ensure efficiency, scalability, and robustness.
Meanwhile, AutoML has emerged as a related area where researchers are automating the construction of these pipelines, and many companies have developed tools for this purpose~\citep{microsoftAzureAutoml,amazonSagemakerAutopilot,googleVertexAI}.
However, to the best of our knowledge, there is no existing studies that systematically discusses the reliability of data analysis pipelines.

In principle, one might attempt to evaluate the reliability of a data analysis pipeline using resampling or data-splitting techniques such as cross-validation.
However, such approaches are often inappropriate for unsupervised learning tasks such as clustering.
In these settings, the analysis itself determines the structures of interest from the data, and the resulting outputs may not follow the same distribution across resampled datasets.
Consequently, the assumption that the data used for evaluation are independent and identically distributed with respect to the selected hypotheses is generally violated, making it difficult to apply resampling-based inference methods in a statistically valid manner.
Additionally, data-splitting approaches reduce the effective sample size available for analysis, which can degrade the accuracy of hypothesis selection and reduce statistical power.
As a recent attempt to enable data splitting for non-i.i.d. data, a technique called data thinning has been proposed~\citep{neufeld2024data,dharamshi2025generalized}, which may be applicable to our clustering pipelines.

In principle, one might attempt to evaluate the reliability of a data analysis pipeline using resampling or data-splitting techniques such as cross-validation.
However, such approaches are often inappropriate for unsupervised learning tasks such as clustering.
In these settings, the analysis itself determines the structures of interest from the data, and the resulting outputs may not follow the same distribution across resampled datasets.
Consequently, the assumption that the data used for evaluation are independent and identically distributed with respect to the selected hypotheses is generally violated, making it difficult to apply resampling-based inference methods in a statistically valid manner.
Additionally, data-splitting approaches reduce the effective sample size available for analysis, which can degrade the accuracy of hypothesis selection and reduce statistical power.
As a recent approach for performing data splitting on non-i.i.d. data, there is a new approach called \emph{data thinning}~\citep{neufeld2024data,dharamshi2025generalized}, which is potentially applicable to our clustering pipelines.

Selective inference (SI) was originally developed for statistical inference after feature selection (FS) in linear models~\citep{taylor2015statistical, lee2016exact}.
Early work studied fundamental FS methods such as the Lasso \citep{lockhart2014significance, lee2016exact} and stepwise feature selection \citep{loftus2014significance}.
The framework was later extended to more complex methods including group Lasso \citep{loftus2015selective}, fused Lasso \citep{hyun2018exact}, and high-order interaction models \citep{suzumura2017selective, das2021fast}.
The key idea of SI is to perform inference conditional on the event that a particular feature selection algorithm selects a given set of variables.
Although conditional inference itself has long been studied in statistics, the work of Lee et al. (2016) provided a practical procedure for computing valid conditional $p$-values and confidence intervals, which established SI as a general framework for inference after data-driven model selection.

SI has been extended beyond feature selection in linear models to a variety of data-driven analysis tasks.
In particular, SI has been applied to unsupervised learning tasks where the hypotheses of interest are selected based on the observed data.
Examples include change-point detection \citep{hyun2016exact, duy2020computing, jewell2022testing, shiraishi2023selective}, outlier detection \citep{chen2020valid, tsukurimichi2021conditional} and anomaly detection \citep{le2024cad, miwa2024statistical}, where SI enables valid inference on structures discovered through the analysis procedure.
In this work, we focus on clustering problems, where a central question is whether clusters identified from data reflect genuine structure or arise from noise.
SI provides a principled framework for addressing this question.
Representative studies include SI methods for hierarchical clustering \citep{gao2022selective} and $k$-means clustering \citep{chen2023selective}.

Another active line of research in SI is devoted for improving statistical power.
One promising direction is randomized selective inference \citep{tian2018selective,panigrahi2024exact,panigrahi2022approximate}, which introduces randomization into the data or the selection procedure to improve statistical power.
Another direction seeks to avoid excessive conditioning in SI that can reduce inferential power.
In this context, line-search--based approaches \citep{le2021parametric, duy2022morepowerful, shiraishi2024bounded} have been proposed to identify minimal conditioning regions while maintaining validity.
Such techniques make it possible to apply SI to complex computational procedures, including deep learning models \citep{duy2022quantifying, miwa2023valid, shiraishi2024statistical, niihori2025quantifying, nishino2025statistical, shiraish2025statistical2}.
Closely related to our work is recent research on SI for feature-selection pipelines \citep{shiraishi2025statistical}, where line-search-based algorithms account for preprocessing steps such as missing-value imputation and anomaly detection.
Inspired by this idea, our work develops SI methods for clustering pipelines operating on complex and heterogeneous data.

\paragraph{Contributions.}
Our contributions in this study are threefold.
First, we develop a statistical test for clustering pipelines composed of various configurations of outlier detection (OD), feature selection (FS), and clustering components, based on the SI framework.
Second, this study represents the first application of SI to inference on a combination of multiple analysis components in clustering contexts in a unified, systematic manner.
Finally, we establish a computational and implementation framework, that facilitates the construction of statistical tests across any clustering pipeline configuration without additional implementation costs.

\newpage
\section{Preliminaries}
\label{sec:preliminaries}

Given a set of algorithm components, a pipeline is defined by selecting some components from the set and connecting the selected components in an appropriate way.
A pipeline can be represented as a directed acyclic graph (DAG) with components as nodes, and the connections as edges.
In this study, as an example class of pipelines, we consider a set of algorithms consisting of two OD algorithms, two FS algorithms, and two clustering algorithms, as well as \emph{Intersection} and \emph{Union} operations (specific algorithms for OD, FS, and clustering are described later in this section).
Figure~\ref{fig:two_examples} shows two examples of pipelines within this class.

\paragraph{Problem Setting.}

In this study, we consider the problem of clustering from a dataset that may contain outliers and/or irrelevant features using the aforementioned class of data analysis pipelines \citep{gao2022selective, chen2023selective}.

Let $X \in \mathbb{R}^{n \times d}$ be the data matrix, where $n$ is the number of samples and $d$ is the number of features.
Since we formulate the probabilistic model in vector form, we also consider the vectorization of $X$, denoted by $\bm{X} \in \mathbb{R}^{nd}$.
Note that $X$ and $\bm{X}$ represent the same underlying data object in matrix and vector forms, respectively.
In the probabilistic formulation below, $\bm{X}$ is treated as a random vector, and the corresponding matrix-form representation $X$ may be regarded as a random matrix by the same convention.

We denote by $\bm{x} \in \mathbb{R}^{nd}$ the observed data, which is assumed to be a realization of $\bm{X}$ generated from the following probabilistic model:
\begin{equation}
    \bm{X} = \bm{\mu} + \bm{\varepsilon}, \quad \bm{\varepsilon} \sim \mathcal{N}(\bm{0}, \bm{\Sigma}),
    \label{eq:statistical_model}
\end{equation}
where $\bm{\mu} \in \mathbb{R}^{nd}$ is the unknown true mean vector, $\bm{\varepsilon} \in \mathbb{R}^{nd}$ is the noise vector, and $\bm{\Sigma}$ is a known covariance matrix.\footnote{The case where $\bm{\Sigma}$ is estimated from the data is discussed in Appendix~\ref{app:estimated_variance}.}
Note that this does not mean that the data themselves follow a Gaussian distribution. Rather, it assumes that Gaussian noise is added to underlying true values that may have an arbitrary structure (for example, being partitioned into multiple clusters).

Using the above notation, a data analysis pipeline comprising OD, FS, and clustering algorithm components is represented as the following function:
\begin{equation}
    \mathcal{P}: \mathbb{R}^{n \times d} \ni X \mapsto (\mathcal{O}, \mathcal{M}, \mathcal{C}) \in 2^{[n]} \times 2^{[d]} \times \{1, \dots, K\}^n,
    \label{eq:pre_pipeline_function}
\end{equation}
where $\mathcal{O} \subset [n] \coloneqq \{1, \dots, n\}$ is the set of detected outlier indices, $\mathcal{M} \subset [d] \coloneqq \{1, \dots, d\}$ is the set of selected feature indices, and $\mathcal{C} \in \{-1, 1, \dots, K\}^n$ is the cluster label vector, with $\mathcal{C}_i = -1$ indicating that sample $i$ is identified as an outlier.

\paragraph{Statistical Test for Pipelines.}

Given the output of a pipeline in~\eqref{eq:pre_pipeline_function}, the statistical significance of the finally obtained cluster structure can be quantified based on the difference in sample means between two clusters, computed only from the dataset after outlier removal and feature selection have been applied.
To formalize this, we denote the data matrix after removing outliers and composed only of the selected features as $X_{(-\mathcal{O}, \mathcal{M})} \in \mathbb{R}^{(n - |\mathcal{O}|) \times |\mathcal{M}|}$, and denote the corresponding submatrix of the true mean matrix as $\mu_{(-\mathcal{O}, \mathcal{M})} \in \mathbb{R}^{(n - |\mathcal{O}|) \times |\mathcal{M}|}$.
Let $\mathcal{C}_a$ and $\mathcal{C}_b$ be the index sets of the s belonging to two clusters of interest.
To test whether there is a statistically significant difference in the true mean values between the two clusters for a selected feature $j \in \mathcal{M}$, we set the null hypothesis $\mathrm{H}_0$ and the alternative hypothesis $\mathrm{H}_1$ as follows:
\begin{equation}
    \begin{aligned}
        \mathrm{H}_0 &: \frac{1}{|\mathcal{C}_a|} \sum_{i \in \mathcal{C}_a} (\mu_{(-\mathcal{O}, \mathcal{M})})_{ij} = \frac{1}{|\mathcal{C}_b|} \sum_{i \in \mathcal{C}_b} (\mu_{(-\mathcal{O}, \mathcal{M})})_{ij}, \\
        &\hspace{9.5em} \text{vs.} \\
        \mathrm{H}_1 &: \frac{1}{|\mathcal{C}_a|} \sum_{i \in \mathcal{C}_a} (\mu_{(-\mathcal{O}, \mathcal{M})})_{ij} \neq \frac{1}{|\mathcal{C}_b|} \sum_{i \in \mathcal{C}_b} (\mu_{(-\mathcal{O}, \mathcal{M})})_{ij}.
    \end{aligned}
    \label{eq:pre_hypothesis}
\end{equation}
The test statistic $T(\bm{X})$ is defined as the difference in sample means between the two clusters:
\begin{equation}
    T(\bm{X}) = \frac{1}{|\mathcal{C}_a|} \sum_{i \in \mathcal{C}_a} (X_{(-\mathcal{O}, \mathcal{M})})_{ij} - \frac{1}{|\mathcal{C}_b|} \sum_{i \in \mathcal{C}_b} (X_{(-\mathcal{O}, \mathcal{M})})_{ij} = \bm{\eta}^\top \bm{X},
    \label{eq:pre_test_statistic}
\end{equation}
where $\bm{\eta} \in \mathbb{R}^{nd}$ is a vector whose $\ell$-th element, where $\ell = (i-1)d + j$ for $i \in [n]$ and $j \in [d]$, is defined as
\begin{equation*}
    \eta_{\ell} =
    \begin{cases}
        \dfrac{1}{|\mathcal{C}_a|} & \text{if } i \in \mathcal{C}_a \text{ and } j \in \mathcal{M}, \\[6pt]
        -\dfrac{1}{|\mathcal{C}_b|} & \text{if } i \in \mathcal{C}_b \text{ and } j \in \mathcal{M}, \\[6pt]
        0 & \text{otherwise},
    \end{cases}
\end{equation*}
where $1\{\cdot\}$ denotes the indicator function.

\paragraph{Outlier Detection (OD) Algorithm Components.}

In this paper, we consider two distance-based outlier detection algorithms as examples of OD algorithms: $k$-NN Removal and $k$-NN-mean Removal (see Appendix~\ref{app:od} for details).
An OD algorithm component is represented as:
\begin{equation*}
    f_{\mathrm{OD}}: (X, \mathcal{O}, \mathcal{M}, \mathcal{C}) \mapsto (X, \mathcal{O}^\prime, \mathcal{M}, \mathcal{C}),
\end{equation*}
where $\mathcal{O}^\prime$ is the updated set of outlier indices.
Note that, if outlier removal has not yet been performed, the set $\mathcal{O}$ is initialized as $\mathcal{O} = \emptyset$.
These algorithms can be applied either as pre-processing before clustering or as post-processing after clustering.\footnote{When applied as pre-processing, outlier detection is performed on the entire dataset. When applied as post-processing, it is performed within each cluster separately.}

\paragraph{Feature Selection (FS) Algorithm Components.}

In this paper, as two examples of FS algorithms, we consider variance- and correlation-based feature selection algorithms (see Appendix~\ref{app:fs} for details).
An FS algorithm component is represented as:
\begin{equation*}
    f_{\mathrm{FS}}: (X, \mathcal{O}, \mathcal{M}, \mathcal{C}) \mapsto (X, \mathcal{O}, \mathcal{M}^\prime, \mathcal{C}),
\end{equation*}
where $\mathcal{M}^\prime$ is the updated set of selected feature indices.
Note that, if feature selection has not yet been performed, the set $\mathcal{M}$ is initialized as $\mathcal{M} = [d]$.

\paragraph{Clustering Algorithm Components.}

In this paper, as two examples of clustering algorithms, we consider DBSCAN and $k$-means (see Appendix~\ref{app:clustering} for details).
A clustering algorithm component is represented as:
\begin{equation*}
    f_{\mathrm{C}}: (X, \mathcal{O}, \mathcal{M}, \mathcal{C}) \mapsto (X, \mathcal{O}, \mathcal{M}, \mathcal{C}^\prime),
\end{equation*}
where $\mathcal{C}^\prime$ is the updated cluster label vector.
The $k$-means clustering component is based on the SI method of \citet{chen2023selective}, extended with a line-search-based approach to improve statistical power.
The DBSCAN clustering component adapts the SI method of \citet{phu2025statistical} to the setting of testing for differences between identified clusters.
Note that, unlike OD and FS, the clustering algorithm is assumed to be applied only once in a pipeline.
Accordingly, if clustering has not yet been performed, the vector $\mathcal{C}$ is initialized as $\mathcal{C} = \mathbf{0}_n$.

\paragraph{Union and Intersection Components.}

When using multiple OD or FS algorithms in combination, it is necessary to include components in the pipeline that integrate the detected outlier sets or selected feature sets via union or intersection operations.
Such integration components for OD and FS are respectively written as:
\begin{gather*}
    f^{\mathcal{O}}_{\Sigma}: (X, \{\mathcal{O}_e\}_{e \in [E]}, \mathcal{M}, \mathcal{C}) \mapsto (X, \Sigma_{e \in [E]} \mathcal{O}_e, \mathcal{M}, \mathcal{C}), \\
    f^{\mathcal{M}}_{\Sigma}: (X, \mathcal{O}, \{\mathcal{M}_e\}_{e \in [E]}, \mathcal{C}) \mapsto (X, \mathcal{O}, \Sigma_{e \in [E]} \mathcal{M}_e, \mathcal{C}),
\end{gather*}
where $E$ is the number of OD/FS algorithms, and the operator $\Sigma$ denotes either the union ($\bigcup$) or intersection ($\bigcap$) of the sets.
Taking the union removes any point (or feature) identified by at least one method, while taking the intersection retains only those identified by all methods.

\newpage
\section{Selective Inference for Clustering Pipelines}
\label{sec:sec3}
To perform statistical tests for clustering pipelines, it is necessary to account for how the data influence the final result through each pipeline component and their composition under a given configuration.
We address this challenge by utilizing the SI framework.
In the SI framework, statistical inference is performed based on the sampling distribution conditional on the process by which the data selects the final result, thereby incorporating the influence of how data is processed in the pipeline.

\paragraph{Selective Inference.}
In SI, $p$-values are computed based on the null distribution conditional on an event that a certain hypothesis is selected.
The goal of SI is to compute a $p$-value such that
\begin{equation}
    \label{eq:conditional_type_i_error_rate}
    \mathbb{P}_{\mathrm{H}_0}
    \left(
    p \leq \alpha \mid
    \mathcal{O}_{\bm{X}} = \mathcal{O}_{\bm{x}},\,
    \mathcal{M}_{\bm{X}} = \mathcal{M}_{\bm{x}},\,
    \mathcal{C}_{\bm{X}} = \mathcal{C}_{\bm{x}}
    \right)
    = \alpha,\ \forall\alpha\in (0,1),
\end{equation}
where $\mathcal{O}_{\bm{X}}$, $\mathcal{M}_{\bm{X}}$, and $\mathcal{C}_{\bm{X}}$ are random variables representing the outlier set, feature set, and clustering result, respectively, derived by applying the clustering pipeline to the random data vector $\bm{X}$.
On the other hand, $\mathcal{O}_{\bm{x}}$, $\mathcal{M}_{\bm{x}}$, and $\mathcal{C}_{\bm{x}}$ denote their specific realizations obtained by applying the pipeline to the observed data $\bm{x}$.
Therefore, the conditioning in~\eqref{eq:conditional_type_i_error_rate} means that we restrict our attention to data $\bm{X}$ that yields the same outlier set $\mathcal{O}_{\bm{x}}$, feature set $\mathcal{M}_{\bm{x}}$, and cluster labels $\mathcal{C}_{\bm{x}}$ as those obtained from the observed data $\bm{x}$.
If the conditional type I error rate can be controlled as in~\eqref{eq:conditional_type_i_error_rate} for all possible pipeline outputs $(\mathcal{O}, \mathcal{M}, \mathcal{C}) \in 2^{[n]} \times 2^{[d]} \times \{1,\dots,K\}^n$, then, by the law of total probability, the marginal type I error rate can also be controlled for all $\alpha \in (0,1)$ because
\begin{align*}
     & \mathbb{P}_{\mathrm{H}_0}(p \leq \alpha) \\
     & = \sum_{\mathcal{O} \in 2^{[n]}}
       \sum_{\mathcal{M} \in 2^{[d]}}
       \sum_{\mathcal{C} \in \{1,\dots,K\}^n}
    \mathbb{P}_{\mathrm{H}_0}(\mathcal{O}, \mathcal{M}, \mathcal{C}) \cdot
    \mathbb{P}_{\mathrm{H}_0}
    \!\left(
    p \leq \alpha \mid
    \substack{
        \mathcal{O}_{\bm{X}} = \mathcal{O}_{\bm{x}},\,
        \mathcal{M}_{\bm{X}} = \mathcal{M}_{\bm{x}},\\
        \mathcal{C}_{\bm{X}} = \mathcal{C}_{\bm{x}}
    }
    \right) \\
     & = \alpha.
\end{align*}
Therefore, in order to perform a valid statistical test, we can employ $p$-values conditional on the pipeline output selection event.
To compute a $p$-value that satisfies~\eqref{eq:conditional_type_i_error_rate}, we need to derive the sampling distribution of the test statistic
\begin{equation}
    \label{eq:conditional_test_statistic}
    T(\bm{X}) \mid
    \left\{
    \mathcal{O}_{\bm{X}} = \mathcal{O}_{\bm{x}},\,
    \mathcal{M}_{\bm{X}} = \mathcal{M}_{\bm{x}},\,
    \mathcal{C}_{\bm{X}} = \mathcal{C}_{\bm{x}}
    \right\}.
\end{equation}

\paragraph{Selective $p$-value.}
To conduct statistical hypothesis testing based on the conditional sampling distribution in~\eqref{eq:conditional_test_statistic}, we introduce an additional condition on the sufficient statistic of the nuisance parameter $\mathcal{Q}_{\bm{X}}$, defined as
\begin{equation}
    \label{eq:nuisance_parameter}
    \mathcal{Q}_{\bm{X}} =
    \left(
    I_{nd} - \frac{\bm{\Sigma}\bm{\eta}\bm{\eta}^{\top}}{\bm{\eta}^{\top}\bm{\Sigma}\bm{\eta}}
    \right)
    \bm{X}.
\end{equation}
This additional conditioning on $\mathcal{Q}_{\bm{X}}$ is a standard practice in the SI literature required for eliminating the nuisance parameters\footnote{
    The nuisance component $\mathcal{Q}_{\bm{X}}$ corresponds to the component $\bm{z}$ in the seminal paper~\citep{lee2016exact} (see Sec. 5, Eq. (5.2), and Theorem 5.2) and is used in almost all the SI-related works that we cited.
}.
Based on the additional conditioning on $\mathcal{Q}_{\bm{X}}$, the following theorem tells that the conditional $p$-value that satisfies~\eqref{eq:conditional_type_i_error_rate} can be derived by using a truncated normal distribution.
\begin{theorem}
    \label{thm:conditional_sampling_distribution}
    Consider a random data vector $\bm{X} \sim \mathcal{N}(\bm{\mu}, \bm{\Sigma})$ and an observed data vector $\bm{x}$.
    Let $(\mathcal{O}_{\bm{X}}, \mathcal{M}_{\bm{X}}, \mathcal{C}_{\bm{X}})$ and $(\mathcal{O}_{\bm{x}}, \mathcal{M}_{\bm{x}}, \mathcal{C}_{\bm{x}})$ be the pipeline outputs obtained by applying a pipeline $\mathcal{P}$ in the form of~\eqref{eq:pre_pipeline_function} to $\bm{X}$ and $\bm{x}$, respectively.
    Let $\bm{\eta} \in \mathbb{R}^{nd}$ be a vector depending on $(\mathcal{O}_{\bm{x}}, \mathcal{M}_{\bm{x}}, \mathcal{C}_{\bm{x}})$, and consider a test statistic in the form of $T(\bm{X}) = \bm{\eta}^{\top}\bm{X}$.
    Furthermore, define the nuisance parameter $\mathcal{Q}_{\bm{X}}$ as in~\eqref{eq:nuisance_parameter}.
    Then, the conditional distribution
    \begin{equation*}
        T(\bm{X}) \mid
        \left\{
        \mathcal{O}_{\bm{X}} = \mathcal{O}_{\bm{x}},\,
        \mathcal{M}_{\bm{X}} = \mathcal{M}_{\bm{x}},\,
        \mathcal{C}_{\bm{X}} = \mathcal{C}_{\bm{x}},\,
        \mathcal{Q}_{\bm{X}} = \mathcal{Q}_{\bm{x}}
        \right\}
    \end{equation*}
    follows a truncated normal distribution $\mathrm{TN}(\bm{\eta}^{\top}\bm{\mu},\, \bm{\eta}^{\top}\bm{\Sigma}\bm{\eta},\, \mathcal{Z})$ with mean $\bm{\eta}^{\top}\bm{\mu}$, variance $\bm{\eta}^{\top}\bm{\Sigma}\bm{\eta}$, and truncation region $\mathcal{Z}$.
    The truncation region $\mathcal{Z}$ is defined as
    \begin{equation}
        \label{eq:truncation_intervals}
        \mathcal{Z} = \left\{
        z \in \mathbb{R} \mid
        \mathcal{O}_{\bm{a}+\bm{b}z} = \mathcal{O}_{\bm{x}},\,
        \mathcal{M}_{\bm{a}+\bm{b}z} = \mathcal{M}_{\bm{x}},\,
        \mathcal{C}_{\bm{a}+\bm{b}z} = \mathcal{C}_{\bm{x}}
        \right\},
    \end{equation}
    \begin{equation*}
        \bm{a} = \mathcal{Q}_{\bm{x}}, \quad
        \bm{b} = \frac{\bm{\Sigma}\bm{\eta}}{\bm{\eta}^{\top}\bm{\Sigma}\bm{\eta}}.
    \end{equation*}
\end{theorem}
The proof of Theorem~\ref{thm:conditional_sampling_distribution} is deferred to Appendix~\ref{app:proof_truncated}.
Based on Theorem~\ref{thm:conditional_sampling_distribution}, we define the selective $p$-value as follows.
Let $\mathcal{X}$ be the conditional data space defined as
\begin{equation*}
    \mathcal{X} = \left\{
    \bm{X} \in \mathbb{R}^{nd} \mid
    {\mathcal{O}}_{\bm{X}} = {\mathcal{O}}_{\bm{x}},
    {\mathcal{M}}_{\bm{X}} = {\mathcal{M}}_{\bm{x}},
    {\mathcal{C}}_{\bm{X}} = {\mathcal{C}}_{\bm{x}},
    {\mathcal{Q}}_{\bm{X}} = {\mathcal{Q}}_{\bm{x}}
    \right\}.
\end{equation*}
We define the pivot quantity $\pi$ and the selective $p$-value $p_{\mathrm{selective}}$ as
\begin{align}
    \pi &\coloneqq \mathbb{P}_{\mathrm{H}_0} \left( T(\bm{X}) \geq T(\bm{x}) \mid \bm{X} \in \mathcal{X} \right), \nonumber\\
    p_{\mathrm{selective}} &\coloneqq 2 \min (\pi, 1 - \pi). \label{eq:p_selective_pivot}
\end{align}
By Theorem~\ref{thm:conditional_sampling_distribution}, the conditional distribution of $T(\bm{X})$ given $\bm{X} \in \mathcal{X}$ follows $\mathrm{TN}(\bm{\eta}^{\top}\bm{\mu},\, \bm{\eta}^{\top}\bm{\Sigma}\bm{\eta},\, \mathcal{Z})$.
Under $\mathrm{H}_0$, this implies $\pi \sim \mathrm{Uniform}[0, 1]$, meaning that $p_{\mathrm{selective}}$ can be computed from the truncated normal distribution once the truncation region $\mathcal{Z}$ in~\eqref{eq:truncation_intervals} is identified.
\begin{theorem}
    \label{thm:property_of_selective_p_value}
    The selective $p$-value defined in~\eqref{eq:p_selective_pivot} satisfies the property in~\eqref{eq:conditional_type_i_error_rate}, i.e.,
    \begin{equation*}
        \label{eq:thm_first_term}
        \mathbb{P}_{\mathrm{H}_0}
        \left(
        p_{\mathrm{selective}} \leq \alpha
        \,\middle|\,
        \begin{gathered}
            \mathcal{O}_{\bm{X}} = \mathcal{O}_{\bm{x}},\\
            \mathcal{M}_{\bm{X}} = \mathcal{M}_{\bm{x}},\\
            \mathcal{C}_{\bm{X}} = \mathcal{C}_{\bm{x}}
        \end{gathered}
        \right)
        = \alpha,\ \forall\alpha \in (0,1).
    \end{equation*}
    Then, the selective $p$-value also satisfies the following property of a valid $p$-value:
    \begin{equation*}
        \mathbb{P}_{\mathrm{H}_0}(p_{\mathrm{selective}} \leq \alpha) = \alpha,\ \forall\alpha \in (0,1).
    \end{equation*}
\end{theorem}
The proof of Theorem~\ref{thm:property_of_selective_p_value} is deferred to Appendix~\ref{app:proof_property_of_selective_p_value}.
This theorem guarantees that the selective $p$-value is uniformly distributed under the null hypothesis $\mathrm{H}_0$, and thus can be used to conduct the valid statistical inference in~\eqref{eq:pre_hypothesis}.
Once the truncation region $\mathcal{Z}$ is identified, the selective $p$-value in~\eqref{eq:p_selective_pivot} can be easily computed by Theorem~\ref{thm:conditional_sampling_distribution}.
Thus, the remaining task is reduced to identifying the truncation region $\mathcal{Z}$.

\newpage
\section{Computations: Line Search Interpretation}
\label{sec:sec4}
From the discussion in~\S\ref{sec:sec3}, it is sufficient to identify the one-dimensional subset $\mathcal{Z}$ in~\eqref{eq:truncation_intervals} to conduct the inference.
In this section, we propose a novel line search method to efficiently identify $\mathcal{Z}$.

\subsection{Overview of the Line Search}
\label{subsec:overview_line_search}
\begin{figure}[H]
    \centering
    \includegraphics[width=0.88\textwidth]{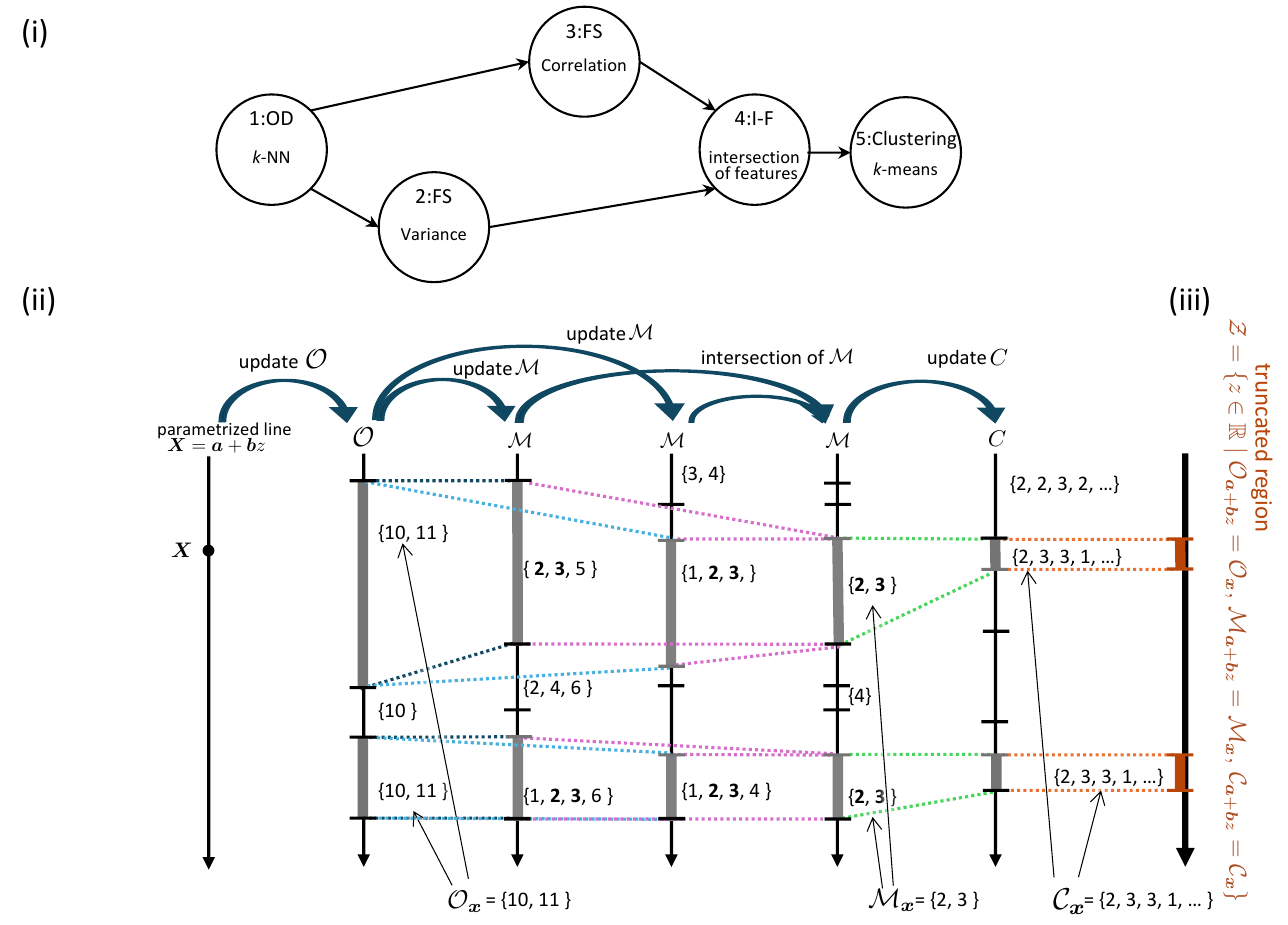}
    \caption[Schematic diagram of the proposed line search method to identify the truncated region $\mathcal{Z}$.]{
        Schematic diagram of the proposed line search method to identify the truncated region $\mathcal{Z}$.
        The top part shows the DAG representation of the pipeline and its topological sorting (i).
        The lower left part shows the operations executed sequentially according to the update rules (ii).
        The lower right part shows how the truncated region $\mathcal{Z}$ is identified by taking the union of several intervals based on parametric-programming (iii).
    }
    \label{fig:linear_search_overview}
\end{figure}

The difficulty in identifying $\mathcal{Z}$ arises from the fact that multiple OD/FS algorithms and a clustering algorithm are applied in an arbitrary complex order.
To surmount this difficulty, we propose an efficient search method that leverages parametric-programming and the fact that our pipeline can be conceptualized as a directed acyclic graph (DAG) whose nodes represent the operations.
In a standard clustering pipeline, $\mathcal{O}$, $\mathcal{M}$, and $\mathcal{C}$ are computed and updated along the DAG.
However, in our framework, intervals for which $\mathcal{O}$, $\mathcal{M}$, and $\mathcal{C}$ are constant are also computed and updated, allowing the computation of the truncation region $\mathcal{Z}$.
In the following, we first discuss how, given a certain computational procedure (combining \emph{update rules} as discussed later), the truncation region $\mathcal{Z}$ can be identified by parametric-programming, and then describe the update rules for each node based on the existing SI methods for each OD, FS, and clustering algorithm.
Since DAGs admit a topological ordering, the update rules can be applied sequentially along the sorted nodes.
The overview of the proposed line search method is illustrated in Figure~\ref{fig:linear_search_overview}.

\subsection{Parametric-Programming}
\label{subsec:parametric_programming}
To identify the truncation region $\mathcal{Z}$, we assume that we have a procedure to compute the interval $[L_z, U_z]$ for any $z \in \mathbb{R}$ which satisfies
\begin{equation*}
    \forall r \in [L_z, U_z], \quad
    \mathcal{O}_{\bm{a}+\bm{b}r} = \mathcal{O}_{\bm{a}+\bm{b}z}, \quad
    \mathcal{M}_{\bm{a}+\bm{b}r} = \mathcal{M}_{\bm{a}+\bm{b}z}, \quad
    \mathcal{C}_{\bm{a}+\bm{b}r} = \mathcal{C}_{\bm{a}+\bm{b}z}.
\end{equation*}
Then, the truncation region $\mathcal{Z}$ can be obtained as the union of the intervals $[L_z, U_z]$ as
\begin{equation}
    \label{eq:pp}
    \mathcal{Z} =
    \bigcup_{
        \substack{
            z \in \mathbb{R} \mid \\
            \mathcal{O}_{\bm{a}+\bm{b}z} = \mathcal{O}_{\bm{x}}, \\
            \mathcal{M}_{\bm{a}+\bm{b}z} = \mathcal{M}_{\bm{x}}, \\
            \mathcal{C}_{\bm{a}+\bm{b}z} = \mathcal{C}_{\bm{x}}
        }
    }
    [L_z, U_z].
\end{equation}
The procedure in~\eqref{eq:pp} is commonly referred to as parametric-programming~\citep{duy2022morepowerful}
We discuss the details of the procedure to compute the interval $[L_z, U_z]$ by defining the update rules for each node in the next subsection.

\subsection{Update Rules}
\label{subsec:update_rules}
In this subsection, we discuss the computational procedure to obtain the interval $[L_z, U_z]$ for any $z \in \mathbb{R}$.
To compute the interval $[L_z, U_z]$, we consider the input of each node in the DAG as a tuple $(\bm{a}, \bm{b}, z, \mathcal{O}, \mathcal{M}, \mathcal{C}, l, u)$ including $\bm{a}, \bm{b} \in \mathbb{R}^{nd}$ and $z \in \mathbb{R}$, where $\mathcal{O}$, $\mathcal{M}$, and $\mathcal{C}$ are the current outlier set, selected feature set, and cluster labels, respectively, and $l, u \in \mathbb{R}$ are the current lower and upper bounds of the interval.
The input of the first node is initialized to $(\bm{a}, \bm{b}, z, \emptyset, [d], \mathbf{0}_n, -\infty, \infty)$.
We detail the update rules for this tuple at each node of the DAG in Appendix~\ref{app:update_rules}.
The overall procedure for computing the interval $[L_z, U_z]$ by applying the update rules in the order of the topological sorting of the DAG is summarized in Algorithm~\ref{alg:auto_conditioning}, where the operation $\mathrm{pa}$ receives the index of the target node and returns the index of its parent node, with $\mathrm{pa}(1)$ defined as $0$.
Algorithm~\ref{alg:auto_conditioning} satisfies the specifications described in~\S\ref{subsec:parametric_programming}, i.e., the following theorem holds.
\begin{theorem}
    \label{thm:auto_conditioning}
    Consider a pipeline $\mathcal{P}$ and vectors $\bm{a}, \bm{b} \in \mathbb{R}^{nd}$ that linearly represent the data.
    For any $z \in \mathbb{R}$, let $[L_z, U_z]$, $\mathcal{O}_{\bm{a}+\bm{b}z}$, $\mathcal{M}_{\bm{a}+\bm{b}z}$, and $\mathcal{C}_{\bm{a}+\bm{b}z}$ be the outputs of Algorithm~\ref{alg:auto_conditioning} with $\mathcal{P}$, $\bm{a}$, $\bm{b}$, and $z$ as inputs.
    Then, for any $r \in [L_z, U_z]$, the output of Algorithm~\ref{alg:auto_conditioning} does not change by changing the input $z$ to $r$:
    \begin{equation*}
        \mathtt{UpdateInterval}(\mathcal{P}, \bm{a}, \bm{b}, r)
        = \bigl([L_z, U_z],\, \mathcal{O}_{\bm{a}+\bm{b}z},\, \mathcal{M}_{\bm{a}+\bm{b}z},\, \mathcal{C}_{\bm{a}+\bm{b}z}\bigr).
    \end{equation*}
\end{theorem}
The proof of Theorem~\ref{thm:auto_conditioning} is deferred to Appendix~\ref{app:proof_auto_conditioning}.
\begin{algorithm}
    \caption{Apply Update Rules in Order of Topological Sorting of DAG (Update Interval)}
    \label{alg:auto_conditioning}
    \begin{algorithmic}[1]
        \renewcommand{\algorithmicrequire}{\textbf{Input:}}
        \renewcommand{\algorithmicensure}{\textbf{Output:}}
        \REQUIRE $\mathcal{P}$, $\bm{a}$, $\bm{b}$ and $z$
        \STATE Convert the pipeline $\mathcal{P}$ to a topologically sorted graph $(V, E)$.
        \STATE Initialize the input of the first node $B_0$ as $(\bm{a}, \bm{b}, z, \emptyset, [d], \mathbf{0}_n, -\infty, \infty)$ (see \S\ref{subsec:update_rules}).
        \FOR{each index of node $i \in \{1, \ldots, |V|\}$}
            \STATE Apply the update rule of the node $v_i$ to its input $B_{\mathrm{pa}(i)}$ to obtain the output $B_i$ (see \S\ref{subsec:update_rules}).
        \ENDFOR
        \STATE Let the components of the last output $B_{|V|}$ corresponding to $\mathcal{O}, \mathcal{M}, \mathcal{C}, l, u$ be $\mathcal{O}_{\bm{a}+\bm{b}z}, \mathcal{M}_{\bm{a}+\bm{b}z}, \mathcal{C}_{\bm{a}+\bm{b}z}, L_z$, and $U_z$, respectively.
        \ENSURE $[L_z, U_z]$, $\mathcal{O}_{\bm{a}+\bm{b}z}$, $\mathcal{M}_{\bm{a}+\bm{b}z}$, and $\mathcal{C}_{\bm{a}+\bm{b}z}$.
    \end{algorithmic}
\end{algorithm}

\newpage
\section{Numerical Experiments}
\label{sec:sec5}
\subsection{Synthetic Data Experiments}
\label{sec:synthetic_data_experiments}

\paragraph{Methods for Comparison.}
In our experiments, we consider the two types of pipelines: \texttt{option1} and \texttt{option2}, whose configurations are shown in Figure~\ref{fig:two_examples}.
For each pipeline, we compare the proposed method (\texttt{proposed}) with \texttt{w/o-pp} (a method that identifies the truncation region as a single interval without parametric-programming), the naive test (\texttt{naive}), and the Bonferroni correction (\texttt{bonferroni}), in terms of Type I error rate and power.
See Appendix~\ref{app:methods_for_comparison} for more details on the methods for comparison.

\paragraph{Experimental Setup.}
In all experiments, we set the significance level $\alpha = 0.05$ and the covariance matrix to $\bm{\Sigma} = I_{nd} \in \mathbb{R}^{nd \times nd}$.
See Appendix~\ref{app:additional_exp_results} for the experiments under a correlated covariance matrix $\bm{\Sigma}_{ij} = (2^{-|i-j|})_{ij} \in \mathbb{R}^{nd \times nd}$.
We also conducted two robustness experiments to evaluate the Type I error rate control of the proposed method: one with the variance estimated from the same data, and another with the noise following non-Gaussian distributions (see Appendix~\ref{app:robustness} for details).

For the experiments on Type I error rate, data generation and testing were performed 10{,}000 times using null datasets with no cluster structure, generated according to the statistical model in~\eqref{eq:statistical_model} with $\bm{\mu} = \bm{0}$, under the following two settings:
\begin{itemize}
    \item Varying the number of instances $n \in \{100, 150, 200, 250\}$ with the number of features fixed at $d = 10$.
    \item Varying the number of features $d \in \{5, 10, 15, 20\}$ with the sample size fixed at $n = 100$.
\end{itemize}

For the power experiments, we generated datasets with $n = 100$ samples, $d = 10$ features, and a three-cluster structure ($K = 3$), with the signal $\Delta$ varied among $\{0.4, 0.6, 0.8\}$.
The datasets were constructed as follows:
\begin{itemize}
    \item \textit{Cluster structure}: Cluster centers were placed in the first three dimensions using uniformly angled positions on a circle of radius $5\Delta$, where $\Delta \in [0, 1]$ controls the inter-cluster distance.
    \item \textit{Feature structure}: Features 4--6 were set to zero (irrelevant features), while features 7--10 were constructed to have strong pairwise linear dependencies: features 7 and 8 share a common factor, as do features 9 and 10 (with opposite sign), each drawn from $\mathcal{N}(0, 2.5^2)$.
    \item \textit{Outliers}: Outliers accounted for approximately 10\% of the total samples ($n_{\mathrm{outlier}} = 10$), divided into two equal groups:
    \begin{itemize}
        \item \emph{Large outliers}: placed far from the data distribution by shifting their mean to $\pm 8$ in random directions across all features.
        \item \emph{Small outliers}: placed near cluster boundaries by shifting their mean $1.5$ units outward from the nearest cluster center along the corresponding axis, while keeping features 7--10 consistent with the inlier distribution.
    \end{itemize}
\end{itemize}

\begin{figure}[H]
    \centering
    {
        \begin{minipage}[b]{0.48\linewidth}
            \centering
            \includegraphics[width=\linewidth]{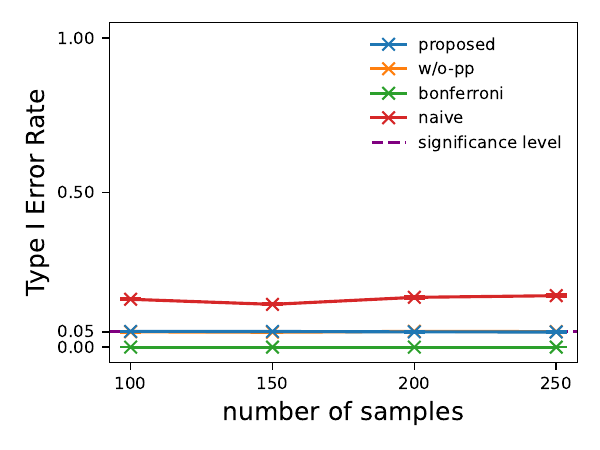}
        \end{minipage}
        \hfill
        \begin{minipage}[b]{0.48\linewidth}
            \centering
            \includegraphics[width=\linewidth]{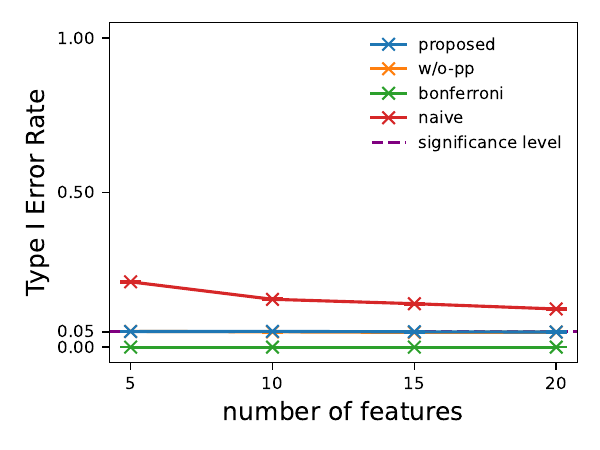}
        \end{minipage}
        \subcaption{Type I Error Rate of \texttt{option1} pipeline}
    }
    \vspace{0.5em}
    {
        \begin{minipage}[b]{0.48\linewidth}
            \centering
            \includegraphics[width=\linewidth]{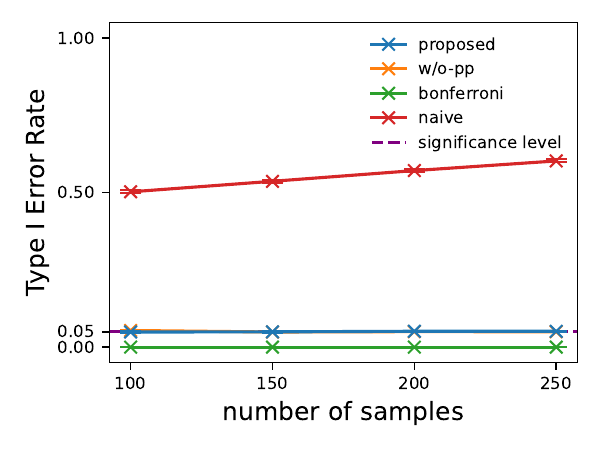}
        \end{minipage}
        \hfill
        \begin{minipage}[b]{0.48\linewidth}
            \centering
            \includegraphics[width=\linewidth]{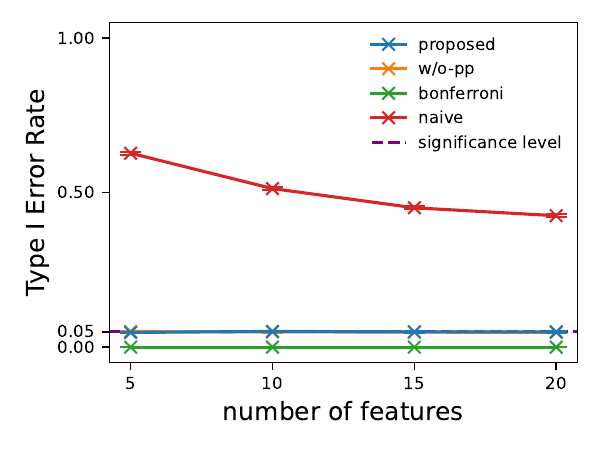}
        \end{minipage}
        \subcaption{Type I Error Rate of \texttt{option2} pipeline}
    }
    \vspace{0.5em}
    {
        \begin{minipage}[b]{0.48\linewidth}
            \centering
            \includegraphics[width=\linewidth]{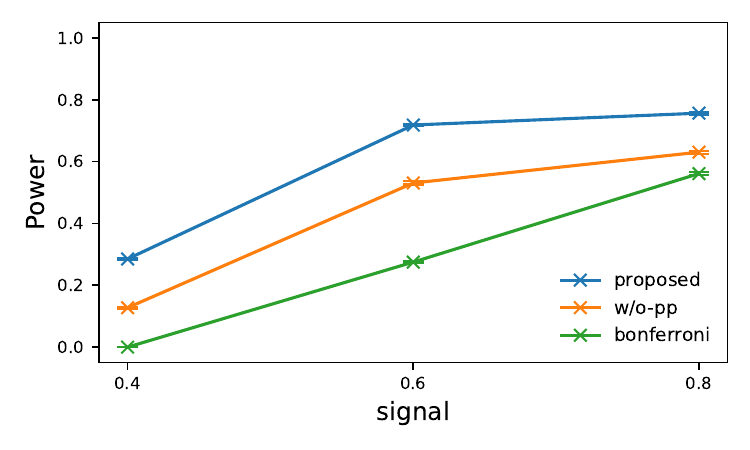}
        \end{minipage}
        \hfill
        \begin{minipage}[b]{0.48\linewidth}
            \centering
            \includegraphics[width=\linewidth]{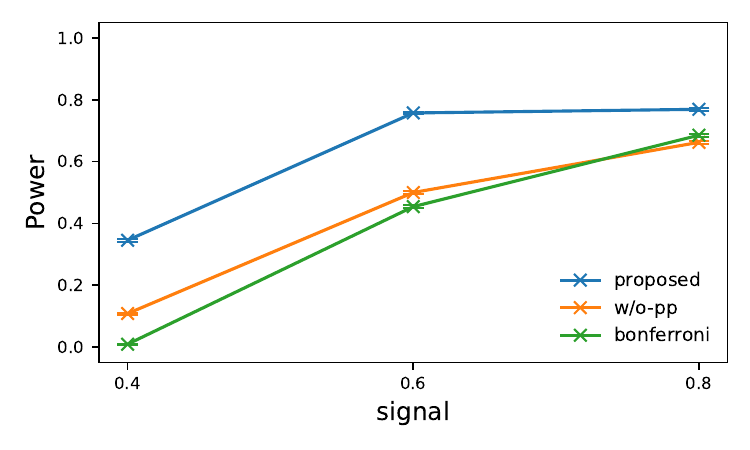}
        \end{minipage}
        \subcaption{Power of \texttt{option1} (left) and \texttt{option2} (right) pipeline}
    }
    \caption{
        Type I error rate when changing the number of samples $n$ and the number of features $d$ (top and middle rows), and power when changing the signal $\Delta$ (bottom row).
        Both \texttt{proposed} and \texttt{w/o-pp} successfully control the Type I error rate, whereas \texttt{naive} fails to do so.
        Among the valid methods, \texttt{proposed} achieves the highest power in all settings for both pipelines.
    }
    \label{fig:main_results}
\end{figure}

\paragraph{Results.}
The results are shown in Figure~\ref{fig:main_results}.
From the Type I error rate results (top and middle rows), \texttt{proposed}, \texttt{w/o-pp}, and \texttt{bonferroni} successfully controlled the Type I error rate below the significance level $\alpha$ in all settings for both pipelines, whereas \texttt{naive} could not.
Since \texttt{naive} failed to control the Type I error rate, its power is not evaluated.
From the power results (bottom row), \texttt{proposed} achieves the highest power among all valid methods.
The power of \texttt{w/o-pp} is reduced by excessive conditioning, while that of \texttt{bonferroni} is reduced by the large number of hypotheses tested.

\clearpage

\subsection{Real Data Experiments}
\label{sec:real_data_experiments}

\paragraph{Target Datasets.}
We consider two real-world datasets:
\begin{itemize}
    \item \textbf{PBMC 3k}~\citep{10x_pbmc3k}: A dataset of approximately 3,000 peripheral blood mononuclear cells (PBMCs), containing diverse immune cell types with approximately 32,000 gene expression features per cell.
    \item \textbf{Marketing Campaign}~\citep{marketing_campaign_dataset}: A dataset of 2,240 customers, each described by 15 features including recency of purchase and spending on product categories.
\end{itemize}
Both datasets are well-suited for clustering tasks aimed at discovering unknown group structures:
identifying patient subgroups with distinct biological characteristics in PBMC 3k, and identifying customer segments with high purchasing potential in Marketing Campaign.

\paragraph{Experimental Setup.}
For both datasets, each feature is log-transformed as $\log(1+x)$.
For PBMC 3k, we additionally use the top 20 highly variable genes, while for Marketing Campaign, all 15 features are used.
To verify the effectiveness of the proposed method, we construct two types of sub-datasets from each dataset:
\begin{itemize}
    \item \textbf{Clustered data}: A sub-dataset with a clear cluster structure. For PBMC 3k, two distinct cell types (CD4 T cells and B cells) are mixed. For Marketing Campaign, customers are randomly drawn from the full dataset to include diverse customer groups.
    \item \textbf{Non-clustered data}: A sub-dataset without a true cluster structure. For PBMC 3k, only a single cell type (CD4 T cells) is used. For Marketing Campaign, only customers with income within median $\pm$ IQR are selected, forming a homogeneous group.
\end{itemize}
For each sub-dataset, 600 samples are extracted in total: 200 for testing and 400 for covariance matrix estimation.
The significance level is set to $\alpha = 0.05$ with hyperparameters fixed per setting.The significance level is set to $\alpha = 0.05$ with hyperparameters fixed per setting.
For each dataset, we apply the clustering pipeline and compare the $p$-values obtained by \texttt{naive} and \texttt{proposed}.

\paragraph{Results.}
Tables~\ref{table:pbmc3k-option1-compare} and~\ref{table:Marketing-option2-compare} show the $p$-values obtained by \texttt{naive} and \texttt{proposed} for the PBMC 3k (with \texttt{option1}) and Marketing Campaign (with \texttt{option2}) datasets, respectively.
In the \textbf{clustered data}, the proposed method yields $p$-values below the significance level for almost all features, successfully detecting the underlying cluster structure.
In the \textbf{non-clustered data}, \texttt{naive} yields $p = 0.000$ for all features due to selection bias, incorrectly detecting a cluster structure.
In contrast, the proposed method yields $p$-values above the significance level for most features, correctly reflecting the absence of a true cluster structure.
See Appendix~\ref{app:real_data_additional_exp_details} for additional results including the \texttt{option2} results for PBMC 3k and the \texttt{option1} results for Marketing Campaign, as well as violin plot visualizations of representative features.

\begin{table}[H]
    \centering
    \caption{Results for clustered and non-clustered data using \texttt{option1} (PBMC 3k).}
    \hspace*{-0.7cm}
    \label{table:pbmc3k-option1-compare}
    \begin{minipage}{0.48\textwidth}
        \centering
        \scriptsize
        \caption*{clustered data (\texttt{option1})}
        \setlength{\tabcolsep}{4pt}
        \begin{tabular}{clcc}
            \toprule
            \# & feature (gene name) & naive & proposed \\
            \midrule
            3  & RPL34  & 0.000 & 0.000 \\
            6  & JUNB   & 0.000 & 0.000 \\
            8  & S100A4 & 0.000 & 0.000 \\
            9  & MT-CO2 & 0.000 & 0.000 \\
            11 & JUN    & 0.000 & 0.000 \\
            18 & LTB    & 0.000 & 0.000 \\
            \bottomrule
        \end{tabular}
    \end{minipage}
    \hspace{0.04\textwidth}
    \begin{minipage}{0.48\textwidth}
        \centering
        \scriptsize
        \caption*{non-clustered data (\texttt{option1})}
        \setlength{\tabcolsep}{4pt}
        \begin{tabular}{clcc}
            \toprule
            \# & feature (gene name) & naive & proposed \\
            \midrule
            0 & RPL34  & 0.000 & 0.052 \\
            1 & JUNB   & 0.000 & 0.261 \\
            2 & S100A4 & 0.000 & 0.604 \\
            3 & MT-CO2 & 0.000 & 0.000 \\
            4 & LTB    & 0.000 & 0.485 \\
            5 & JUN    & 0.000 & 0.000 \\
            \bottomrule
        \end{tabular}
    \end{minipage}
    \hspace*{-0.7cm}
\end{table}

\begin{table}[H]
    \centering
    \caption{Results for clustered and non-clustered data using \texttt{option2} (Marketing Campaign).}
    \hspace*{-1.8cm}
    \label{table:Marketing-option2-compare}
    \begin{minipage}{0.49\textwidth}
        \centering
        \scriptsize
        \caption*{clustered data (\texttt{option2})}
        \setlength{\tabcolsep}{3pt}
        \begin{tabular}{clcc}
            \toprule
            \# & feature (spending target etc.) & naive & proposed \\
            \midrule
            3 & Recency          & 0.000 & 0.000 \\
            5 & MntFruits        & 0.000 & 0.021 \\
            7 & MntFishProducts  & 0.000 & 0.012 \\
            8 & MntSweetProducts & 0.000 & 0.024 \\
            9 & MntGoldProds     & 0.000 & 0.000 \\
            \bottomrule
        \end{tabular}
    \end{minipage}
    \hspace{0.08\textwidth}
    \begin{minipage}{0.49\textwidth}
        \centering
        \scriptsize
        \caption*{non-clustered data (\texttt{option2})}
        \setlength{\tabcolsep}{3pt}
        \begin{tabular}{clcc}
            \toprule
            \# & feature (spending target etc.) & naive & proposed \\
            \midrule
            3 & Recency          & 0.000 & 0.000 \\
            4 & MntWines         & 0.000 & 0.395 \\
            5 & MntFruits        & 0.000 & 0.611 \\
            6 & MntMeatProducts  & 0.000 & 0.482 \\
            8 & MntSweetProducts & 0.000 & 0.645 \\
            9 & MntGoldProds     & 0.000 & 0.387 \\
            \bottomrule
        \end{tabular}
    \end{minipage}
    \hspace*{-1.2cm}
\end{table}
\

\newpage
\section{Conclusion}
\label{sec:sec6}
In this study, we developed a selective-inference framework for clustering pipelines composed of multiple OD, FS, and clustering components, and established exact control of the Type I error rate for arbitrary configurations within the considered class.
This work represents a first step toward valid statistical inference for clustering procedures including preprocessing, ensuring the reliability of identified cluster structures.
Our long-term goal is to ensure the reproducibility of data-driven decision-making by accounting for the entire analysis pipeline from raw data to final conclusions, with the present study serving as a proof of concept in the clustering setting.
Future work includes extending the class of admissible components, incorporating model selection procedures such as cross-validation, and developing computational tools that enable practical deployment of pipeline-aware selective inference.

\newpage
\subsubsection*{Acknowledgments}
This work was partially supported by JST CREST (JPMJCR21D3, JPMJCR22N2), JST Moonshot R\&D (JPMJMS2033-05), and RIKEN Center for Advanced Intelligence Project.

\clearpage
\bibliographystyle{plainnat}
\bibliography{ref}

\clearpage
\appendix

\newpage
\section{Pipeline Components}
\label{app:pipeline_components}
\begin{figure}[H]
    \centering
    \includegraphics[width=1.0\linewidth]{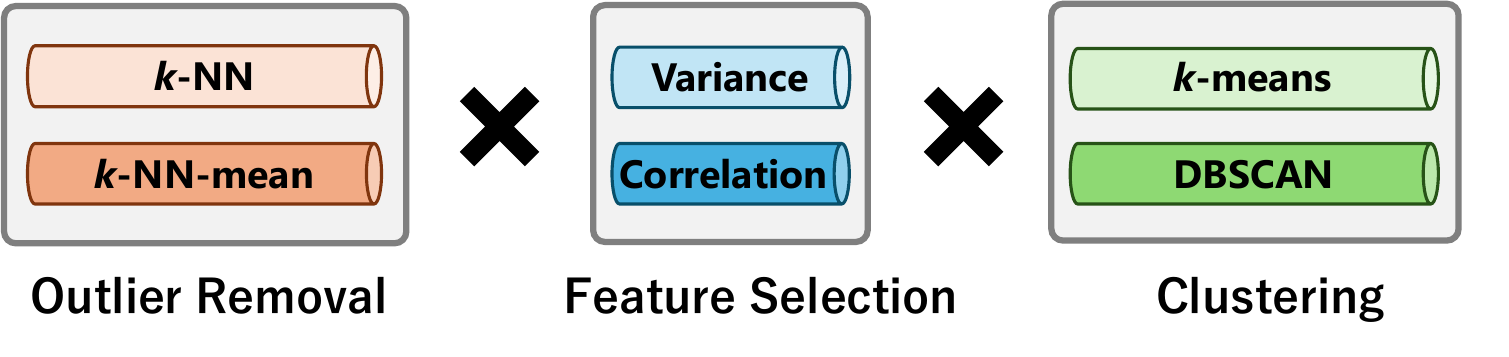}
    \caption{Algorithm components of the clustering pipeline.}
    \label{fig:clustering_pipeline_algorithms}
\end{figure}

In this study, as a demonstration of the clustering pipeline framework, we adopt two OD algorithms, two FS algorithms, and two clustering algorithms, as illustrated in Figure~\ref{fig:clustering_pipeline_algorithms}.
This appendix describes the details of each algorithm component and the set aggregation operations.
As noted in \S\ref{sec:preliminaries}, each algorithm in the pipeline is applied to the submatrix $X_{(-\mathcal{O}, \mathcal{M})} \in \mathbb{R}^{(n-|\mathcal{O}|) \times |\mathcal{M}|}$ obtained by applying the results $(\mathcal{O}, \mathcal{M})$ of the preceding steps; however, for brevity of notation, the input data to each algorithm is generally denoted as $X \in \mathbb{R}^{n \times d}$ throughout this appendix.

\subsection{Outlier Detection (OD) Algorithm Components}
\label{app:od}
An OD algorithm component is represented as
\begin{equation*}
    f_{\mathrm{OD}}: ( X, \mathcal{O}, \mathcal{M}, \mathcal{C} ) \mapsto ( X, \mathcal{O}^{\prime}, \mathcal{M}, \mathcal{C} ),
\end{equation*}
where $\mathcal{O}^{\prime}$ is the updated set of outlier indices, and the set $\mathcal{O}$ is initialized as $\mathcal{O} = \emptyset$.
OD algorithms can be applied as a preprocessing step or a postprocessing step after clustering.

\paragraph{$k$-NN Removal.}
\label{app:knn_removal}
$k$-Nearest Neighbor Removal ($k$-NN Removal) identifies isolated points that deviate significantly from the data distribution as outliers, based on the distance between each data point and its neighbors.
For each data point $\bm{X}_i$ in $X = \{\bm{X}_1, \dots, \bm{X}_n\}^\top \in \mathbb{R}^{n \times d}$, we compute an anomaly score by searching for its $k$-nearest neighbors.
When applied as a postprocessing step after clustering, the search range is restricted to data points belonging to the same cluster as $\bm{X}_i$.
Let $\bm{X}_i^{(k)}$ denote the data point with the $k$-th smallest Euclidean distance from $\bm{X}_i$ among all candidates excluding itself.
The anomaly score $d_k(\bm{X}_i)$ is defined as
\begin{equation*}
    d_k(\bm{X}_i) = \| \bm{X}_i - \bm{X}_i^{(k)} \|_2^2.
\end{equation*}
For a predefined threshold $\tau$, the data point $\bm{X}_i$ is identified as an outlier if $d_k(\bm{X}_i) > \tau$, and the updated outlier index set is obtained as
\begin{equation*}
    \mathcal{O}^{\prime} = \mathcal{O} \cup \{ i \in \{1, \dots, n\} \mid d_k(\bm{X}_i) > \tau \}.
\end{equation*}

\paragraph{$k$-NN-mean Removal.}
\label{app:knn_average_removal}
$k$-Nearest Neighbor Average Removal ($k$-NN-mean Removal) is a variant of $k$-NN Removal that uses the average distance to the $k$ nearest neighbors, rather than the distance to the single $k$-th nearest neighbor, as the anomaly score.
For a data point $\bm{X}_i$, the anomaly score $\bar{d}_k(\bm{X}_i)$ is defined as
\begin{equation*}
    \bar{d}_k(\bm{X}_i) = \frac{1}{k} \sum_{j=1}^{k} \| \bm{X}_i - \bm{X}_i^{(j)} \|_2^2.
\end{equation*}
The updated outlier index set $\mathcal{O}^{\prime}$ is obtained in the same manner as in $k$-NN Removal, by applying the threshold $\tau$ to $\bar{d}_k(\bm{X}_i)$.
Compared to $k$-NN Removal, this method provides a smoother reflection of the local density around each data point, offering improved robustness against noise at the cost of sensitivity to extreme outliers.

\subsection{Feature Selection (FS) Algorithm Components}
\label{app:fs}
An FS algorithm component is represented as
\begin{equation*}
    f_{\mathrm{FS}}: ( X, \mathcal{O}, \mathcal{M}, \mathcal{C} ) \mapsto ( X, \mathcal{O}, \mathcal{M}^{\prime}, \mathcal{C} ),
\end{equation*}
where $\mathcal{M}^{\prime}$ is the updated set of selected feature indices, and the set $\mathcal{M}$ is initialized as $\mathcal{M} = \{1, \dots, d\}$.
FS algorithms are applied as a preprocessing step before clustering, typically after OD.

\paragraph{Variance-based Feature Selection.}
\label{app:variance_based_fs}
This method evaluates the variability (variance) of each feature and removes low-variance features that are considered to carry little information.
For each feature $j \in \{1, \dots, d\}$, the sample variance $\hat{\sigma}_j^2$ is computed as
\begin{equation*}
    \hat{\sigma}_j^2 = \frac{1}{n-1} \sum_{i=1}^{n} (X_{ij} - \bar{X}_j)^2, \quad \text{where} \quad \bar{X}_j = \frac{1}{n} \sum_{i=1}^{n} X_{ij}.
\end{equation*}
The updated feature index set $\mathcal{M}^{\prime}$ is then obtained as
\begin{equation*}
    \mathcal{M}^{\prime} = \{ j \in \mathcal{M} \mid \hat{\sigma}_j^2 > \tau \},
\end{equation*}
where $\tau$ is a predefined threshold.
This processing removes features with nearly constant values or near-noise components, retaining only features likely to carry meaningful structure for clustering.

\paragraph{Correlation-based Feature Selection.}
\label{app:correlation_based_fs}
This method removes redundant features by eliminating one feature from each pair of highly correlated features, thereby reducing multicollinearity and improving clustering stability.
For any two distinct features $j, k \in \mathcal{M}$, the sample correlation coefficient $\hat{\rho}_{jk}$ is computed as
\begin{equation*}
    \hat{\rho}_{jk} = \frac{\hat{\sigma}_{jk}}{\sqrt{\hat{\sigma}_j^2 \hat{\sigma}_k^2}}, \quad
    \hat{\sigma}_{jk} = \frac{1}{n-1} \sum_{i=1}^{n} (X_{ij} - \bar{X}_j)(X_{ik} - \bar{X}_k).
\end{equation*}
For a predefined correlation threshold $\tau_{\mathrm{corr}}$, the feature with the larger index in each highly correlated pair is regarded as redundant, and its index set is defined as
\begin{equation*}
    R = \{ j \in \mathcal{M} \mid \exists\, k < j \ \text{s.t.}\ |\hat{\rho}_{jk}| > \tau_{\mathrm{corr}} \}.
\end{equation*}
The updated feature index set is then obtained as
\begin{equation*}
    \mathcal{M}^{\prime} = \mathcal{M} \setminus R.
\end{equation*}

\subsection{Clustering Algorithm Components}
\label{app:clustering}
A clustering algorithm component is represented as
\begin{equation*}
    f_{\mathrm{C}}: ( X, \mathcal{O}, \mathcal{M}, \mathcal{C} ) \mapsto ( X, \mathcal{O}, \mathcal{M}, \mathcal{C}^{\prime} ),
\end{equation*}
where $\mathcal{C}^{\prime}$ is the updated set of cluster labels, and $\mathcal{C}$ is initialized as $\mathcal{C} = \mathbf{0}_n$.

\paragraph{$k$-means Clustering.}
\label{app:kmeans_clustering}
$k$-means clustering partitions a given dataset into $K$ clusters by solving the following optimization problem:
\begin{equation*}
    \operatorname*{minimize}_{\mathcal{C}_1, \dots, \mathcal{C}_K}
    \left\{
    \sum_{k=1}^{K} \sum_{i \in \mathcal{C}_k}
    \left\| \bm{X}_i - \frac{1}{|\mathcal{C}_k|}\sum_{j \in \mathcal{C}_k} \bm{X}_j \right\|_2^2
    \right\},
    \label{eq:k-means}
\end{equation*}
\begin{equation*}
    \text{subject to} \quad \bigcup_{k=1}^{K} \mathcal{C}_k = \{1, \dots, n\}, \quad \mathcal{C}_k \cap \mathcal{C}_{k^{\prime}} = \emptyset \quad \forall k \neq k^{\prime}.
\end{equation*}
Since finding a globally optimal solution to~\eqref{eq:k-means} is generally intractable, we employ Lloyd's algorithm~\citep{lloyd1982least,macqueen1967some} (Algorithm~\ref{alg:lloyd}) to obtain a locally optimal solution.
\begin{algorithm}[H]
    \caption{Lloyd's algorithm for $k$-means clustering}
    \label{alg:lloyd}
    \begin{algorithmic}[1]
        \renewcommand{\algorithmicrequire}{\textbf{Input:}}
        \renewcommand{\algorithmicensure}{\textbf{Output:}}
        \REQUIRE Data $X = \{\bm{X}_1, \dots, \bm{X}_n\}^\top \in \mathbb{R}^{n \times d}$, number of clusters $K$, maximum iterations $T$, random seed $s$.
        \ENSURE Cluster labels $\mathcal{C}^{\prime} = \{c_1, \dots, c_n\}$.
        \STATE Initialize centroids $(\bm{m}_1^{(0)}, \dots, \bm{m}_K^{(0)})$ by sampling $K$ samples from $\bm{X}_1, \dots, \bm{X}_n$ without replacement using seed $s$.
        \STATE Compute assignments $c_i^{(0)} \leftarrow \operatorname*{argmin}_{1 \le k \le K} \| \bm{X}_i - \bm{m}_k^{(0)} \|_2^2$, $\quad i = 1, \dots, n$.
        \STATE Initialize $t = 0$.
        \WHILE{$t \le T$}
            \STATE Update centroids: $\bm{m}_k^{(t+1)} \leftarrow \left(\sum_{i: c_i^{(t)}=k} \bm{X}_i\right) \big/ \sum_{i=1}^{n} \mathbf{1}\{c_i^{(t)}=k\}$, $\quad k = 1, \dots, K$.
            \STATE Update assignments: $c_i^{(t+1)} \leftarrow \operatorname*{argmin}_{1 \le k \le K} \| \bm{X}_i - \bm{m}_k^{(t+1)} \|_2^2$, $\quad i = 1, \dots, n$.
            \IF{$c_i^{(t+1)} = c_i^{(t)}$ for all $i$}
                \STATE \textbf{break}.
            \ELSE
                \STATE $t \leftarrow t + 1$.
            \ENDIF
        \ENDWHILE
        \RETURN $(c_1^{(t)}, \dots, c_n^{(t)})$.
    \end{algorithmic}
\end{algorithm}

\paragraph{DBSCAN Clustering.}
\label{app:dbscan_clustering}
DBSCAN (Density-Based Spatial Clustering of Applications with Noise)~\citep{ester1996density} is a density-based clustering method that does not require the number of clusters $K$ to be specified in advance and is capable of identifying outliers as noise points.
DBSCAN uses two parameters: a neighborhood radius $\epsilon$ and a minimum number of points $n_{\mathrm{min}}$ required to form a cluster.
Each data point is classified into one of the following three types:
\begin{itemize}
    \item \textbf{Core point}: A point with at least $n_{\mathrm{min}}$ points (including itself) within radius $\epsilon$.
    \item \textbf{Border point}: A point that is within radius $\epsilon$ of a core point but does not itself have $n_{\mathrm{min}}$ neighbors.
    \item \textbf{Noise point}: A point that is neither a core point nor a border point, and does not belong to any cluster.
\end{itemize}
The procedure is summarized in Algorithm~\ref{alg:dbscan_simple} (main routine) and Algorithm~\ref{alg:expand_cluster} (subroutine).
The algorithm iterates over unvisited points, assigns a new cluster to each core point, and expands the cluster by recursively adding density-reachable points.
Points that remain unassigned are labeled as noise ($c_i = -1$).
Unlike $k$-means, DBSCAN is able to correctly extract clusters with nonlinear shapes.

\begin{algorithm}[H]
    \caption{DBSCAN}
    \label{alg:dbscan_simple}
    \begin{algorithmic}[1]
        \renewcommand{\algorithmicrequire}{\textbf{Input:}}
        \renewcommand{\algorithmicensure}{\textbf{Output:}}
        \REQUIRE Data $X = \{\bm{X}_1, \dots, \bm{X}_n\}^\top \in \mathbb{R}^{n \times d}$, radius $\epsilon$, minimum points $n_{\mathrm{min}}$.
        \ENSURE Cluster labels $\mathcal{C}^{\prime} = \{c_1, \dots, c_n\}$ ($c_i = -1$ if noise).
        \STATE Initialize all points as \textbf{unvisited}; set $c_i \leftarrow 0$ for all $i$; set $k \leftarrow 0$.
        \FOR{$i = 1$ to $n$}
            \IF{$\bm{X}_i$ is visited} \STATE \textbf{continue}. \ENDIF
            \STATE Mark $\bm{X}_i$ as visited.
            \STATE $N_i \leftarrow \{ \bm{X}_j \mid \| \bm{X}_i - \bm{X}_j \|_2 \le \epsilon \}$.
            \IF{$|N_i| < n_{\mathrm{min}}$}
                \STATE $c_i \leftarrow -1$ (noise).
            \ELSE
                \STATE $k \leftarrow k + 1$; \textbf{ExpandCluster}$(\bm{X}_i, N_i, k)$.
            \ENDIF
        \ENDFOR
    \end{algorithmic}
\end{algorithm}

\begin{algorithm}[H]
    \caption{ExpandCluster}
    \label{alg:expand_cluster}
    \begin{algorithmic}[1]
        \REQUIRE Core point $\bm{X}_i$, neighbor set $N_i$, cluster index $k$.
        \STATE $c_i \leftarrow k$; initialize queue $S \leftarrow N_i \setminus \{\bm{X}_i\}$.
        \WHILE{$S$ is not empty}
            \STATE Take $\bm{X}_p$ from $S$.
            \IF{$\bm{X}_p$ is not visited}
                \STATE Mark $\bm{X}_p$ as visited.
                \STATE $N_p \leftarrow \{ \bm{X}_j \mid \| \bm{X}_p - \bm{X}_j \|_2 \le \epsilon \}$.
                \IF{$|N_p| \ge n_{\mathrm{min}}$} \STATE $S \leftarrow S \cup N_p$. \ENDIF
            \ENDIF
            \IF{$c_p = 0$ or $c_p = -1$} \STATE $c_p \leftarrow k$. \ENDIF
        \ENDWHILE
    \end{algorithmic}
\end{algorithm}

\newpage
\section{Proofs}
\label{app:proofs}
\subsection{Proof of Theorem~\ref{thm:conditional_sampling_distribution}}
\label{app:proof_truncated}
According to the conditioning on $\mathcal{Q}_{\bm{X}}=\mathcal{Q}_{\bm{x}}$, we have
\begin{equation*}
    \mathcal{Q}_{\bm{X}} = \mathcal{Q}_{\bm{x}} \Leftrightarrow
    \left(
    I_{nd} -
    \frac{\bm{\Sigma}\bm{\eta}\bm{\eta}^{\top}}{\bm{\eta}^{\top}\bm{\Sigma}\bm{\eta}}
    \right)\bm{X} = \mathcal{Q}_{\bm{x}}
    \Leftrightarrow
    \bm{X} = \bm{a} + \bm{b}z,
\end{equation*}
where $z = T(\bm{X}) \in \mathbb{R}$,
$\bm{a} = \mathcal{Q}_{\bm{x}}$, and
$\bm{b} = \bm{\Sigma}\bm{\eta} / (\bm{\eta}^{\top}\bm{\Sigma}\bm{\eta})$
as defined in Theorem~\ref{thm:conditional_sampling_distribution}.
Then, we have
\begin{align*}
      &
    \left\{
    \bm{X}\in\mathbb{R}^{nd}\mid
    \mathcal{O}_{\bm{X}} = \mathcal{O}_{\bm{x}},\,
    \mathcal{M}_{\bm{X}} = \mathcal{M}_{\bm{x}},\,
    \mathcal{C}_{\bm{X}} = \mathcal{C}_{\bm{x}},\,
    \mathcal{Q}_{\bm{X}} = \mathcal{Q}_{\bm{x}}
    \right\}  \\
    = &
    \left\{
    \bm{X}\in\mathbb{R}^{nd}\mid
    \mathcal{O}_{\bm{X}} = \mathcal{O}_{\bm{x}},\,
    \mathcal{M}_{\bm{X}} = \mathcal{M}_{\bm{x}},\,
    \mathcal{C}_{\bm{X}} = \mathcal{C}_{\bm{x}},\,
    \bm{X} = \bm{a} + \bm{b}z,\ z\in\mathbb{R}
    \right\}  \\
    = &
    \left\{
    \bm{a} + \bm{b}z\in\mathbb{R}^{nd}\mid
    \mathcal{O}_{\bm{a}+\bm{b}z} = \mathcal{O}_{\bm{x}},\,
    \mathcal{M}_{\bm{a}+\bm{b}z} = \mathcal{M}_{\bm{x}},\,
    \mathcal{C}_{\bm{a}+\bm{b}z} = \mathcal{C}_{\bm{x}},\,
    z\in\mathbb{R}
    \right\}  \\
    = &
    \left\{
    \bm{a} + \bm{b}z\in\mathbb{R}^{nd}\mid
    z\in \mathcal{Z}
    \right\},
\end{align*}
where $\mathcal{Z}$ is the truncation region defined in~\eqref{eq:truncation_intervals}.
Therefore, we obtain
\begin{equation*}
    T(\bm{X}) \mid
    \left\{
    \mathcal{O}_{\bm{X}} = \mathcal{O}_{\bm{x}},\,
    \mathcal{M}_{\bm{X}} = \mathcal{M}_{\bm{x}},\,
    \mathcal{C}_{\bm{X}} = \mathcal{C}_{\bm{x}},\,
    \mathcal{Q}_{\bm{X}} = \mathcal{Q}_{\bm{x}}
    \right\}
    \sim
    \mathrm{TN}(\bm{\eta}^{\top}\bm{\mu},\, \bm{\eta}^{\top}\bm{\Sigma}\bm{\eta},\, \mathcal{Z}).
\end{equation*}

\subsection{Proof of Theorem~\ref{thm:property_of_selective_p_value}}
\label{app:proof_property_of_selective_p_value}
By probability integral transformation, under the null hypothesis $\mathrm{H}_0$, we have
\begin{equation*}
    p_\mathrm{selective} \mid
    \left\{
    \mathcal{O}_{\bm{X}} = \mathcal{O}_{\bm{x}},\,
    \mathcal{M}_{\bm{X}} = \mathcal{M}_{\bm{x}},\,
    \mathcal{C}_{\bm{X}} = \mathcal{C}_{\bm{x}},\,
    \mathcal{Q}_{\bm{X}} = \mathcal{Q}_{\bm{x}}
    \right\}
    \sim
    \mathrm{Unif}(0, 1),
\end{equation*}
which leads to
\begin{equation*}
    \mathbb{P}_{\mathrm{H}_0}
    \left(
    p_\mathrm{selective} \leq \alpha \mid
    \mathcal{O}_{\bm{X}} = \mathcal{O}_{\bm{x}},\,
    \mathcal{M}_{\bm{X}} = \mathcal{M}_{\bm{x}},\,
    \mathcal{C}_{\bm{X}} = \mathcal{C}_{\bm{x}},\,
    \mathcal{Q}_{\bm{X}} = \mathcal{Q}_{\bm{x}}
    \right)
    =\alpha,\
    \forall\alpha\in(0,1).
\end{equation*}
For any $\alpha\in(0,1)$, by marginalizing over all the values of the nuisance parameters $\mathcal{Q}_{\bm{x}}$, we obtain
\begin{align*}
      &
    \mathbb{P}_{\mathrm{H}_0}
    \left(
    p_\mathrm{selective} \leq \alpha \mid
    \mathcal{O}_{\bm{X}} = \mathcal{O}_{\bm{x}},\,
    \mathcal{M}_{\bm{X}} = \mathcal{M}_{\bm{x}},\,
    \mathcal{C}_{\bm{X}} = \mathcal{C}_{\bm{x}}
    \right)                                                                                                       \\
    = &
    \begin{multlined}
        \int_{\mathbb{R}^{nd}}
        \mathbb{P}_{\mathrm{H}_0}
        \left(
        p_\mathrm{selective} \leq \alpha \mid
        \mathcal{O}_{\bm{X}} = \mathcal{O}_{\bm{x}},\,
        \mathcal{M}_{\bm{X}} = \mathcal{M}_{\bm{x}},\,
        \mathcal{C}_{\bm{X}} = \mathcal{C}_{\bm{x}},\,
        \mathcal{Q}_{\bm{X}} = \mathcal{Q}_{\bm{x}}
        \right) \\
        \times\,\mathbb{P}_{\mathrm{H}_0}
        \left(
        \mathcal{Q}_{\bm{X}} = \mathcal{Q}_{\bm{x}} \mid
        \mathcal{O}_{\bm{X}} = \mathcal{O}_{\bm{x}},\,
        \mathcal{M}_{\bm{X}} = \mathcal{M}_{\bm{x}},\,
        \mathcal{C}_{\bm{X}} = \mathcal{C}_{\bm{x}}
        \right)
        d\mathcal{Q}_{\bm{x}}
    \end{multlined} \\
    = & \;\alpha \int_{\mathbb{R}^{nd}}
    \mathbb{P}_{\mathrm{H}_0}
    \left(
    \mathcal{Q}_{\bm{X}} = \mathcal{Q}_{\bm{x}} \mid
    \mathcal{O}_{\bm{X}} = \mathcal{O}_{\bm{x}},\,
    \mathcal{M}_{\bm{X}} = \mathcal{M}_{\bm{x}},\,
    \mathcal{C}_{\bm{X}} = \mathcal{C}_{\bm{x}}
    \right)
    d\mathcal{Q}_{\bm{x}} = \alpha.
\end{align*}
Therefore, we also obtain
\begin{align*}
      &
    \mathbb{P}_{\mathrm{H}_0}(p_{\mathrm{selective}}\leq \alpha) \\
    = &
    \sum_{\mathcal{O}_{\bm{x}}\in 2^{[n]}}
    \sum_{\mathcal{M}_{\bm{x}}\in 2^{[d]}}
    \sum_{\mathcal{C}_{\bm{x}}\in \{1,\dots,K\}^n}
    \mathbb{P}_{\mathrm{H}_0}(\mathcal{O}_{\bm{x}},\, \mathcal{M}_{\bm{x}},\, \mathcal{C}_{\bm{x}}) \cdot
    \mathbb{P}_{\mathrm{H}_0}
    \!\left(
    p_\mathrm{selective} \leq \alpha \mid
    \substack{
        \mathcal{O}_{\bm{X}} = \mathcal{O}_{\bm{x}},\,
        \mathcal{M}_{\bm{X}} = \mathcal{M}_{\bm{x}},\\
        \mathcal{C}_{\bm{X}} = \mathcal{C}_{\bm{x}}
    }
    \right) \\
    = &
    \;\alpha
    \sum_{\mathcal{O}_{\bm{x}}\in 2^{[n]}}
    \sum_{\mathcal{M}_{\bm{x}}\in 2^{[d]}}
    \sum_{\mathcal{C}_{\bm{x}}\in \{1,\dots,K\}^n}
    \mathbb{P}_{\mathrm{H}_0}(\mathcal{O}_{\bm{x}},\, \mathcal{M}_{\bm{x}},\, \mathcal{C}_{\bm{x}}) = \alpha.
\end{align*}

\subsection{Proof of Theorem~\ref{thm:auto_conditioning}}
\label{app:proof_auto_conditioning}
It is sufficient to consider only $z$ as input to Algorithm~\ref{alg:auto_conditioning}.
In addition, as a notation, we define $\mathcal{G}_i$ as the mapping that returns the last five components of $B_i$ for $i\in \{0,1,\ldots, |V|\}$, i.e.,
\begin{equation*}
    \begin{aligned}
        \mathcal{G}_i\colon
        \mathbb{R}\ni z \mapsto\,
        &\bigl(\mathcal{O}_{\bm{a}+\bm{b}z}^i,\, \mathcal{M}_{\bm{a}+\bm{b}z}^i,\, \mathcal{C}_{\bm{a}+\bm{b}z}^i,\, l_z^i,\, u_z^i\bigr) \\
        &\in 2^{[n]}\times 2^{[d]}\times \{1,\dots,K\}^n \times \mathbb{R}^2,
        \quad i\in \{0, 1, \ldots, |V|\}.
    \end{aligned}
\end{equation*}
According to the above notation, all we have to show is that $\mathcal{G}_{|V|}(z) = \mathcal{G}_{|V|}(r)$ for any $z\in \mathbb{R}$ and any $r\in[l_z^{|V|}, u_z^{|V|}]$.
We show this by mathematical induction.

In the case $i=0$, it is obvious from the initialization in Algorithm~\ref{alg:auto_conditioning} that
$\mathcal{G}_{0}(z) = \mathcal{G}_{0}(r) = (\emptyset,\, [d],\, \mathbf{0}_n,\, -\infty,\, \infty)$
for any $z\in \mathbb{R}$ and any $r\in[l_z^{0}, u_z^{0}]=(-\infty, \infty)$.

Next, we assume that for any fixed $i\in\{0,\ldots,|V|-1\}$, $\mathcal{G}_{j}(z) = \mathcal{G}_{j}(r)$ for any $j\in \{0,\ldots, i\}$, any $z\in \mathbb{R}$ and any $r\in[l_z^{j}, u_z^{j}]$.
Under this assumption, noting that $\mathrm{pa}(i+1)\subset \{0,\ldots, i\}$ from a property of topological sort, it is obvious that $\mathcal{G}_{i+1}(z) = \mathcal{G}_{i+1}(r)$ for any $z\in \mathbb{R}$ and any $r\in[l_z^{i+1}, u_z^{i+1}]$ from the update rule of $v_{i+1}$ described in~\S\ref{subsec:update_rules}.

\newpage
\section{Details of the Update Rules}
\label{app:update_rules}
For each algorithm component, we identify an interval $[l_z, u_z] \ni z$ such that the output of the algorithm (i.e., the outlier set $\mathcal{O}$, the selected feature set $\mathcal{M}$, or the cluster labels $\mathcal{C}$) remains invariant for all $r \in [l_z, u_z]$.
The current interval $[l, u]$ is then updated as $[l^{\prime}, u^{\prime}] = [l_z, u_z] \cap [l, u]$.
This condition is referred to as the \emph{selection event}.
In the following, we describe the identification of $[l, u]$ for each algorithm component.

\subsection{Update Rules for the OD Node.}
The OD Node detects the outliers $\mathcal{O}^{\prime}(z)$ from the dataset $(X_{(-\mathcal{O}, \mathcal{M})},\ \bm{a}_{-\mathcal{O}}+\bm{b}_{-\mathcal{O}}z)$, which means that outlier detection is performed on the submatrix extracted from $(X, \bm{a}+\bm{b}z)$ based on the current $\mathcal{O}$ and $\mathcal{M}$.
For all OD algorithms considered in this study ($k$-NN Removal and $k$-NN-mean Removal; see Appendix~\ref{app:od}), the computation procedure to obtain the interval $[l_z, u_z]\ni z$, which satisfies
\begin{equation*}
    \forall r \in [l_z, u_z], \quad \mathcal{O}^{\prime}(r) = \mathcal{O}^{\prime}(z),
\end{equation*}
has been derived in the latter part of this section.
\begin{equation*}
    \label{eq:update_rule_od}
    (\bm{a}, \bm{b}, z, \mathcal{O}, \mathcal{M}, \mathcal{C}, l, u)
    \mapsto
    (\bm{a}, \bm{b}, z,\ \mathcal{O} \cup \mathcal{O}^{\prime}(z),\ \mathcal{M},\ \mathcal{C},\
    \max(l, l_z),\ \min(u, u_z)).
\end{equation*}
In the following, for notational convenience, we treat the submatrix $X_{(-\mathcal{O}, \mathcal{M})}$ simply as $X \in \mathbb{R}^{n^{\prime} \times d^{\prime}}$, where $n^{\prime} = n - |\mathcal{O}|$ and $d^{\prime} = |\mathcal{M}|$, and its vectorization as $\bm{X} \in \mathbb{R}^{n^{\prime} d^{\prime}}$, in the same manner as in \S\ref{sec:preliminaries}.

\paragraph{$k$-NN Removal.}
From Appendix~\ref{app:knn_removal}, the selection event for $k$-NN Removal consists of the following two conditions:
\begin{enumerate}
    \item \textbf{Preservation of neighbor ordering}: For every data point $\bm{X}_i$, the index $\mathrm{idx}_k(i)$ of the $k$-th nearest neighbor observed at $\bm{x}$ remains the $k$-th nearest neighbor of $\bm{X}_i(z)$.
    \item \textbf{Preservation of threshold decisions}: For each point identified as an outlier, the $k$-NN distance exceeds $\tau$; for each inlier, it does not.
\end{enumerate}
Let $\mathcal{O}^{\mathrm{obs}}$ and $\mathcal{I}^{\mathrm{obs}}$ denote the outlier and inlier sets at the observation $\bm{x}$, respectively.
The neighbor-ordering condition is expressed as the following quadratic inequalities,
which must hold for all $i \in [n^{\prime}]$:
\begin{equation*}
    \| \bm{X}_i(z) - \bm{X}_{\mathrm{idx}_k(i)}(z) \|_2^2 \le \| \bm{X}_i(z) - \bm{X}_j(z) \|_2^2,
    \quad \forall j \in [n^{\prime}] \setminus \mathcal{K}_i,
\end{equation*}
\begin{equation*}
    \| \bm{X}_i(z) - \bm{X}_l(z) \|_2^2 \le \| \bm{X}_i(z) - \bm{X}_{\mathrm{idx}_k(i)}(z) \|_2^2,
    \quad \forall l \in \mathcal{K}_i,
\end{equation*}
where $\mathcal{K}_i$ is the set of $k$-nearest-neighbor indices of $\bm{X}_i$ at $\bm{x}$, computed within $X \in \mathbb{R}^{n^{\prime} \times d^{\prime}}$.
The threshold condition is expressed as
\begin{equation*}
    \begin{cases}
        \| \bm{X}_i(z) - \bm{X}_{\mathrm{idx}_k(i)}(z) \|_2^2 > \tau   & \text{for } i \in \mathcal{O}^{\mathrm{obs}}, \\
        \| \bm{X}_i(z) - \bm{X}_{\mathrm{idx}_k(i)}(z) \|_2^2 \le \tau & \text{for } i \in \mathcal{I}^{\mathrm{obs}}.
    \end{cases}
\end{equation*}
Substituting $\bm{X}(z) = \bm{a} + \bm{b}z$ and letting $\bm{\delta}_{ij}(z) = (\bm{a}_i - \bm{a}_{\mathrm{idx}_j(i)}) + (\bm{b}_i - \bm{b}_{\mathrm{idx}_j(i)})z$ denote the difference vector between $\bm{X}_i(z)$ and its $j$-th nearest neighbor $\bm{X}_{\mathrm{idx}_j(i)}(z)$, each condition reduces to a quadratic inequality of the form $Az^2 + Bz + C \gtrless 0$, which can be solved analytically.
Denoting by $\mathcal{Z}_i^{\mathrm{neighbor}}$ and $\mathcal{Z}_i^{\mathrm{threshold}}$ the intervals derived from these two conditions for each point $i$, the final interval is
\begin{equation*}
    [l_z, u_z] = \bigcap_{i=1}^{n^{\prime}} \left( \mathcal{Z}_i^{\mathrm{neighbor}} \cap \mathcal{Z}_i^{\mathrm{threshold}} \right).
\end{equation*}
Within this interval, $k$-NN Removal is guaranteed to produce exactly the same outlier set as at the observation $\bm{x}$.

\paragraph{$k$-NN-mean Removal.}
The selection event for $k$-NN-mean Removal has the same structure as that for $k$-NN Removal, except that the threshold condition is based on the \emph{average} squared $k$-NN distance rather than the single $k$-th squared distance.
The neighbor-ordering condition is identical to that of $k$-NN Removal, so the interval $\mathcal{Z}_i^{\mathrm{neighbor}}$ is reused directly.
The threshold condition becomes
\begin{equation*}
    \begin{cases}
        \dfrac{1}{k}\displaystyle\sum_{j=1}^{k} \| \bm{X}_i(z) - \bm{X}_{\mathrm{idx}_j(i)}(z) \|_2^2 > \tau   & \text{for } i \in \mathcal{O}^{\mathrm{obs}}, \\[6pt]
        \dfrac{1}{k}\displaystyle\sum_{j=1}^{k} \| \bm{X}_i(z) - \bm{X}_{\mathrm{idx}_j(i)}(z) \|_2^2 \le \tau & \text{for } i \in \mathcal{I}^{\mathrm{obs}}.
    \end{cases}
\end{equation*}
Since the sum of quadratic expressions in $z$ is again quadratic, this condition also reduces to $A_i z^2 + B_i z + C_i \gtrless 0$, and the interval $\mathcal{Z}_i^{\mathrm{mean\_threshold}}$ is obtained analytically.
The final interval is
\begin{equation*}
    [l_z, u_z] = \bigcap_{i=1}^{n^{\prime}} \left( \mathcal{Z}_i^{\mathrm{neighbor}} \cap \mathcal{Z}_i^{\mathrm{mean\_threshold}} \right).
\end{equation*}

\subsection{Update Rules for the FS Node.}
The FS Node selects features $\mathcal{M}^{\prime}(z)$ from the dataset $(X_{(-\mathcal{O}, \mathcal{M})},\ \bm{a}_{-\mathcal{O}}+\bm{b}_{-\mathcal{O}}z)$, which means that feature selection is performed on the submatrix extracted from $(X, \bm{a}+\bm{b}z)$ based on the current $\mathcal{O}$ and $\mathcal{M}$.
For all FS algorithms considered in this study (Variance-based FS and Correlation-based FS; see Appendix~\ref{app:fs}), the computation procedure to obtain the interval $[l_z, u_z]\ni z$, which satisfies
\begin{equation*}
    \forall r \in [l_z, u_z], \quad \mathcal{M}^{\prime}(r) = \mathcal{M}^{\prime}(z),
\end{equation*}
has been derived in the latter part of this section.
Utilizing this, the update rule should be as follows:
\begin{equation*}
    \label{eq:update_rule_fs}
    (\bm{a}, \bm{b}, z, \mathcal{O}, \mathcal{M}, \mathcal{C}, l, u)
    \mapsto
    (\bm{a}, \bm{b}, z,\ \mathcal{O},\ \mathcal{M} \cap \mathcal{M}^{\prime}(z),\ \mathcal{C},\
    \max(l, l_z),\ \min(u, u_z)).
\end{equation*}
In the following, for notational convenience, we treat the submatrix $X_{(-\mathcal{O}, \mathcal{M})}$ simply as $X \in \mathbb{R}^{n^{\prime} \times d^{\prime}}$, where $n^{\prime} = n - |\mathcal{O}|$ and $d^{\prime} = |\mathcal{M}|$, and its vectorization as $\bm{X} \in \mathbb{R}^{n^{\prime} d^{\prime}}$.

\paragraph{Variance-based Feature Selection.}
From Appendix~\ref{app:variance_based_fs}, the selection event consists of the following two conditions:
\begin{enumerate}
    \item \textbf{Preservation of selected features}: For each feature $j \in \mathcal{M}^{\mathrm{obs}}$ selected at $\bm{x}$, the sample variance satisfies $\hat{\sigma}_j^2(z) > \tau$.
    \item \textbf{Preservation of rejected features}: For each feature $j \notin \mathcal{M}^{\mathrm{obs}}$ not selected at $\bm{x}$, the sample variance satisfies $\hat{\sigma}_j^2(z) \le \tau$.
\end{enumerate}
Substituting $X_{ij}(z) = a_{ij} + b_{ij}z$ into the sample variance, the mean $\bar{X}_j(z)$ becomes a linear function of $z$, hence
\begin{equation*}
    \hat{\sigma}_j^2(z) = \mathrm{Var}(\bm{b}_{\cdot j})\, z^2 + 2\,\mathrm{Cov}(\bm{a}_{\cdot j}, \bm{b}_{\cdot j})\, z + \mathrm{Var}(\bm{a}_{\cdot j}),
\end{equation*}
which is a quadratic function of $z$.
Each condition $\hat{\sigma}_j^2(z) \gtrless \tau$ is thus a quadratic inequality $A_j z^2 + B_j z + C_j \gtrless 0$, which can be solved analytically.
Denoting by $\mathcal{Z}_j$ the interval satisfying this condition for each feature $j$, the final interval is
\begin{equation*}
    [l_z, u_z] = \bigcap_{j=1}^{d^{\prime}} \mathcal{Z}_j.
\end{equation*}

\paragraph{Correlation-based Feature Selection.}
From Appendix~\ref{app:correlation_based_fs}, the selection event consists of the following two conditions:
\begin{enumerate}
    \item \textbf{Preservation of removed features}: For each removed feature $j$ and the feature $i < j$ paired with it at $\bm{x}$, the absolute correlation satisfies $|\hat{\rho}_{ij}(z)| > \tau_{\mathrm{corr}}$.
    \item \textbf{Preservation of retained pairs}: For each pair $(i, j)$ deemed uncorrelated at $\bm{x}$, the absolute correlation satisfies $|\hat{\rho}_{ij}(z)| \le \tau_{\mathrm{corr}}$.
\end{enumerate}
Since the covariance $\hat{\sigma}_{ij}(z)$ and the variances $\hat{\sigma}_i^2(z)$, $\hat{\sigma}_j^2(z)$ are all quadratic polynomials in $z$, squaring the condition $|\hat{\rho}_{ij}(z)| \gtrless \tau_{\mathrm{corr}}$ and clearing the denominator yields
\begin{equation*}
    \hat{\sigma}_{ij}(z)^2 \gtrless \tau_{\mathrm{corr}}^2 \cdot \hat{\sigma}_i^2(z) \cdot \hat{\sigma}_j^2(z).
\end{equation*}
The left-hand side is the square of a quadratic in $z$ (hence degree four), and the right-hand side is a product of two quadratics (also degree four), so the constraint reduces to a quartic polynomial inequality $P(z) \gtrless 0$, which can be solved analytically.
Denoting by $\mathcal{Z}_{ij}$ the interval satisfying this condition for each pair $(i, j)$, the final interval is
\begin{equation*}
    [l_z, u_z] = \bigcap_{(i,j) \in \text{All Pairs}} \mathcal{Z}_{ij}.
\end{equation*}

\subsection{Update Rules for the Clustering Node.}
The Clustering Node assigns cluster labels $\mathcal{C}^{\prime}(z)$ to the dataset $X_{(-\mathcal{O}, \mathcal{M})}$ based on the current $\mathcal{O}$ and $\mathcal{M}$.
For all clustering algorithms considered in this study ($k$-means and DBSCAN; see Appendix~\ref{app:clustering}), the computation procedure to obtain the interval $[l_z, u_z]\ni z$, which satisfies
\begin{equation*}
    \forall r \in [l_z, u_z], \quad \mathcal{C}^{\prime}(r) = \mathcal{C}^{\prime}(z),
\end{equation*}
has been derived in the latter part of this section.
Utilizing this, the update rule for a clustering node is given as follows:
\begin{equation*}
    \label{eq:update_rule_clustering}
    (\bm{a}, \bm{b}, z, \mathcal{O}, \mathcal{M}, \mathcal{C}, l, u)
    \mapsto
    (\bm{a}, \bm{b}, z,\ \mathcal{O},\ \mathcal{M},\ \mathcal{C}^{\prime}(z),\
    \max(l, l_z),\ \min(u, u_z)).
\end{equation*}
In the following, for notational convenience, we treat the submatrix $X_{(-\mathcal{O}, \mathcal{M})}$ simply as $X \in \mathbb{R}^{n^{\prime} \times d^{\prime}}$, where $n^{\prime} = n - |\mathcal{O}|$ and $d^{\prime} = |\mathcal{M}|$, and its vectorization as $\bm{X} \in \mathbb{R}^{n^{\prime} d^{\prime}}$.

\paragraph{$k$-means Clustering.}
The update rule for $k$-means clustering follows the approach of \citet{chen2023selective}.
From Appendix~\ref{app:kmeans_clustering}, the selection events in $k$-means clustering are the cluster assignments at each iteration: the initial assignment (Step 2) and the re-assignment after centroid updates (Step 6).
In this study, the random seed used for the initial cluster assignment is fixed at the observed value $\bm{x}$, so that the initial assignment $c_i^{(0)}$ is treated as a deterministic function of $z$ via the data $\bm{X}(z)$.
Let $c_i^{(t)}$ denote the cluster label of $\bm{X}_i$ at iteration $t$ under observation $\bm{x}$.
Since each centroid $m_k^{(t)}$ is a linear function of $\bm{X}$, it is also linear in $z$: $m_k^{(t)}(z) = \bm{a}_{m_k} + \bm{b}_{m_k}z$.
The assignment condition $c_i^{(t)}(z) = c_i^{(t)}(\bm{x})$ is equivalent to
\begin{equation*}
    \left\| \bm{X}_i(z) - m_{c_i^{(t)}}^{(t)}(z) \right\|_2^2 \le \left\| \bm{X}_i(z) - m_k^{(t)}(z) \right\|_2^2, \quad \forall k \ne c_i^{(t)},
\end{equation*}
which reduces to a quadratic inequality in $z$ since both $\bm{X}_i(z)$ and $m_k^{(t)}(z)$ are linear in $z$.
Collecting these conditions over all data points and all iterations defines the intervals $\mathcal{Z}_0$ (initial assignment) and $\mathcal{Z}_T$ (all subsequent assignments), and the final interval is
\begin{equation*}
    [l_z, u_z] = \mathcal{Z}_0 \cap \mathcal{Z}_T.
\end{equation*}
Within this interval, the entire trajectory of Lloyd's algorithm is guaranteed to reproduce the same cluster assignments as at the observation $\bm{x}$.

\paragraph{DBSCAN Clustering.}
The update rule for DBSCAN clustering follows the approach of \citet{phu2025statistical}.
From Appendix~\ref{app:dbscan_clustering}, the cluster structure of DBSCAN is uniquely determined by the $\epsilon$-neighborhood structure of the data.
Therefore, the selection event for DBSCAN is that the neighborhood set $\mathcal{N}_i(\bm{X}(z))$ of every data point $\bm{X}_i$ coincides with the observed neighborhood set $\mathcal{N}_i^{\mathrm{obs}}$, which preserves the core-point decisions and density-reachability relations, and hence the entire clustering result.
This is equivalent to the following two conditions holding simultaneously for all pairs $(i,j)$:
\begin{enumerate}
    \item For each pair that was neighboring at $\bm{x}$: $\|\bm{X}_i(z) - \bm{X}_j(z)\|_2^2 \le \epsilon^2$.
    \item For each pair that was not neighboring at $\bm{x}$: $\|\bm{X}_i(z) - \bm{X}_j(z)\|_2^2 > \epsilon^2$.
\end{enumerate}
Substituting $\bm{X}(z) = \bm{a} + \bm{b}z$ and decomposing the difference vector as $\bm{\delta}_{ij}(z) = (\bm{a}_i - \bm{a}_j) + (\bm{b}_i - \bm{b}_j)z$, each condition reduces to
\begin{equation*}
    \|\bm{b}_i - \bm{b}_j\|_2^2\, z^2 + 2(\bm{a}_i - \bm{a}_j)^\top(\bm{b}_i - \bm{b}_j)\, z + \|\bm{a}_i - \bm{a}_j\|_2^2 - \epsilon^2 \le 0
    \quad (\text{or} > 0),
\end{equation*}
which is a quadratic inequality in $z$.
Denoting by $\mathcal{Z}_{ij}^{\le}$ and $\mathcal{Z}_{ij}^{>}$ the intervals satisfying the neighboring and non-neighboring conditions, respectively, the final interval is
\begin{equation*}
    [l_z, u_z] = \bigcap_{i=1}^{n^{\prime}} \left(
    \bigcap_{j \in \mathcal{N}_i^{\mathrm{obs}}} \mathcal{Z}_{ij}^{\le}
    \;\cap\;
    \bigcap_{j \in [n^{\prime}] \setminus \mathcal{N}_i^{\mathrm{obs}},\, j \ne i} \mathcal{Z}_{ij}^{>}
    \right).
\end{equation*}
Within this interval, the DBSCAN algorithm is guaranteed to produce exactly the same clustering result as at the observation $\bm{x}$.

\subsection{Update Rules for the Node of Union/Intersection.}
The node computes the union or intersection of detected outlier sets or selected feature sets output by $E$ parallel branches.
With $E$ being the number of input edges, for each selected feature and detected outlier, the update rules should be as follows:
\begin{align*}
    \{(\bm{a}, \bm{b}, z, \mathcal{O}_e, \mathcal{M}, \mathcal{C}, l_e, u_e)\}_{e \in [E]}
    &\mapsto \\
    &\hspace{-4em} \Bigl(\bm{a}, \bm{b}, z,\ \textstyle\sum_{e \in [E]} \mathcal{O}_e,\ \mathcal{M},\ \mathcal{C},\
    \max_{e \in [E]} l_e,\ \min_{e \in [E]} u_e\Bigr), \\
    \{(\bm{a}, \bm{b}, z, \mathcal{O}, \mathcal{M}_e, \mathcal{C}, l_e, u_e)\}_{e \in [E]}
    &\mapsto \\
    &\hspace{-4em} \Bigl(\bm{a}, \bm{b}, z,\ \mathcal{O},\ \textstyle\sum_{e \in [E]} \mathcal{M}_e,\ \mathcal{C},\
    \max_{e \in [E]} l_e,\ \min_{e \in [E]} u_e\Bigr),
\end{align*}
where $\sum$ represents the union or intersection depending on the type of the node.

\newpage
\section{Details of the Experiments}
\subsection{Methods for Comparison}
\label{app:methods_for_comparison}
We compared our proposed method with the following methods:
\begin{itemize}
    \item \texttt{w/o-pp}: Our proposed method conditioning on the only one interval $[L_z, U_z]$ to which the observed test statistic $T(\bm{x})$ belongs. This
          method is computationally efficient, however, its power is low due to over-conditioning.
    \item \texttt{naive}: This method uses a classical $z$-test without conditioning, i.e., we compute the naive $p$-value as $p_\mathrm{naive}=\mathbb{P}_{\mathrm{H}_0}(|T(\bm{X})|\geq |T(\bm{x})|)$.
    \item \texttt{Bonferroni}: A Bonferroni correction-based method accounting for the $3^n \cdot 2^d$ possible pipeline outputs (the $3^n$ possible cluster assignments of $n$ data points into two clusters or neither, and the $2^d$ feature selection combinations), where the $p$-value is computed as
          \begin{equation*}
              p_{\mathrm{bonferroni}} = \min\!\left(1,\; 3^n \cdot 2^d \cdot p_{\mathrm{naive}}\right).
          \end{equation*}
\end{itemize}
\subsection{Additional Experiments Results}
\label{app:additional_exp_results}
\begin{figure}[H]
    \centering
    {
        \begin{minipage}[b]{0.48\linewidth}
            \centering
            \includegraphics[width=\linewidth]{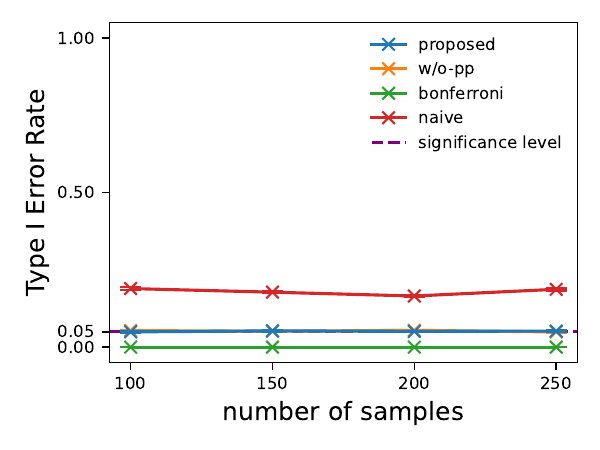}
        \end{minipage}
        \hfill
        \begin{minipage}[b]{0.48\linewidth}
            \centering
            \includegraphics[width=\linewidth]{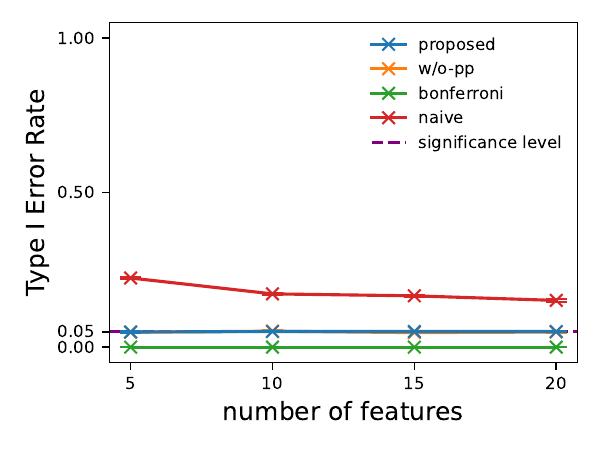}
        \end{minipage}
        \subcaption{Type I Error Rate of \texttt{option1} pipeline}
    }
    \vspace{0.5em}
    {
        \begin{minipage}[b]{0.48\linewidth}
            \centering
            \includegraphics[width=\linewidth]{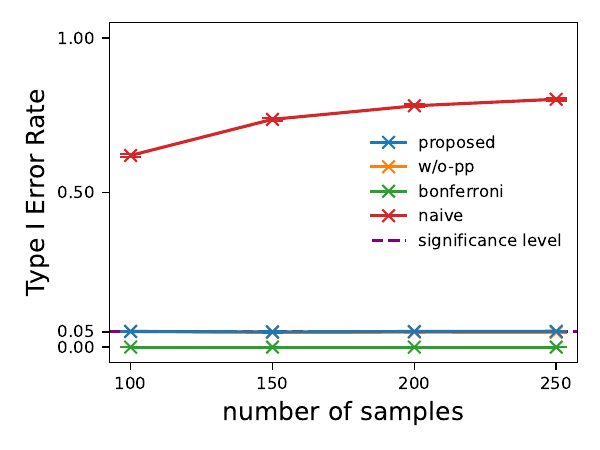}
        \end{minipage}
        \hfill
        \begin{minipage}[b]{0.48\linewidth}
            \centering
            \includegraphics[width=\linewidth]{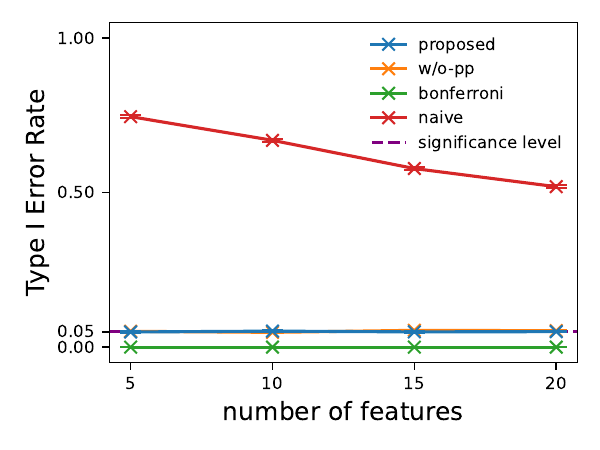}
        \end{minipage}
        \subcaption{Type I Error Rate of \texttt{option2} pipeline}
    }
    \vspace{0.5em}
    {
        \begin{minipage}[b]{0.48\linewidth}
            \centering
            \includegraphics[width=\linewidth]{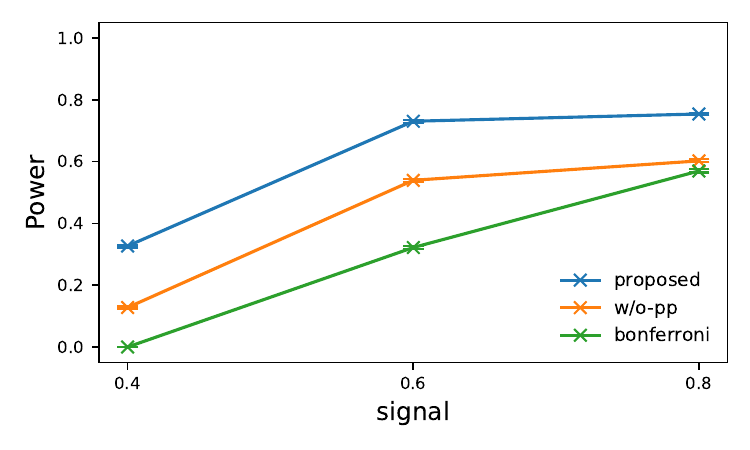}
        \end{minipage}
        \hfill
        \begin{minipage}[b]{0.48\linewidth}
            \centering
            \includegraphics[width=\linewidth]{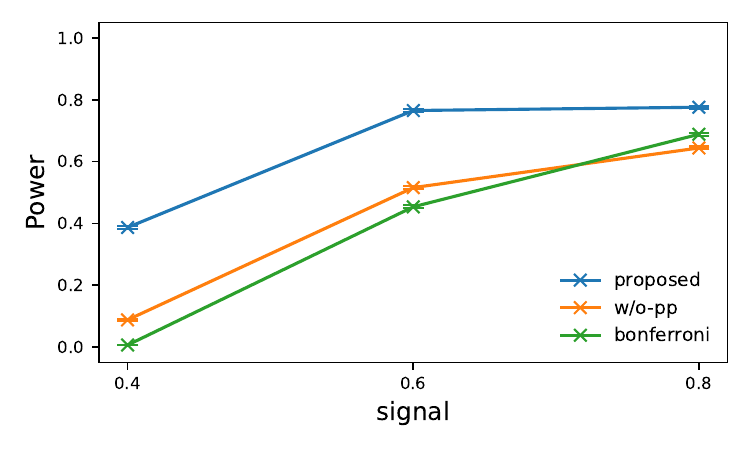}
        \end{minipage}
        \subcaption{Power of \texttt{option1} (left) and \texttt{option2} (right) pipeline}
    }
    \caption{
        Results under correlated covariance matrix $\bm{\Sigma}_{ij} = (2^{-|i-j|})_{ij}$.
        Type I error rate when changing the number of samples $n$ and the number of features $d$ (top and middle) for each pipeline, and power when changing the signal $\Delta$ (bottom).
        Both \texttt{proposed} and \texttt{w/o-pp} successfully control the Type I error rate, whereas \texttt{naive} fails to do so.
        Among the valid methods, \texttt{proposed} achieves the highest power in all settings for both pipelines.
    }
    \label{fig:corr_results}
\end{figure}

We also conducted experiments under a correlated covariance matrix $\bm{\Sigma}_{ij} = (2^{-|i-j|})_{ij} \in \mathbb{R}^{nd \times nd}$ to evaluate the robustness of the proposed method.
The datasets were generated in the same way as in the main experiments (\S\ref{sec:sec5}), and the results are shown in Figure~\ref{fig:corr_results}.
\subsection{Details of the Real Data Experiments}
\label{app:real_data_additional_exp_details}
\subsubsection{PBMC 3k Data}
\paragraph{Additional Results for \texttt{option2}.}
Tables~\ref{table:pbmc3k-notdif-compare-app} shows the \texttt{option2} results for the clustered and non-clustered data, respectively (see \S\ref{sec:sec5} for the dataset and setup details).
The overall trend is consistent with the \texttt{option1} results in \S\ref{sec:sec5}: the proposed method successfully detects the cluster structure in the clustered data, and correctly suppresses false positives in the non-clustered data.
In particular, \texttt{option2} yields high $p$-values for most features in the non-clustered data, demonstrating appropriate decision-making.
For features \#2 (S100A4) and \#4 (LTB), the proposed method still yields $p$-values below the significance level; since the Type I error rate is properly controlled, these results suggest the presence of genuine within-cell-type heterogeneity such as individual differences or cell-cycle variation.
\begin{table}[H]
    \centering
    \caption{Results of \texttt{option2} for clustered and non-clustered data (PBMC 3k).}
    \hspace*{-0.7cm}
    \label{table:pbmc3k-dif-compare-app}
    \begin{minipage}{0.48\textwidth}
        \centering
        \scriptsize
        \caption*{clustered data (\texttt{option2})}
        \setlength{\tabcolsep}{4pt}
        \begin{tabular}{clcc}
            \toprule
            \# & feature (gene name) & naive & proposed (SI) \\
            \midrule
            3  & RPL34  & 0.000 & 0.000 \\
            6  & JUNB   & 0.000 & 0.000 \\
            8  & S100A4 & 0.000 & 0.000 \\
            10 & FOS    & 0.000 & 0.000 \\
            11 & JUN    & 0.000 & 0.000 \\
            14 & VIM    & 0.000 & 0.000 \\
            15 & IL32   & 0.000 & 0.000 \\
            16 & DUSP1  & 0.000 & 0.000 \\
            17 & ACTG1  & 0.000 & 0.009 \\
            18 & LTB    & 0.000 & 0.000 \\
            \bottomrule
        \end{tabular}
    \end{minipage}
    \hspace{0.04\textwidth}
    \begin{minipage}{0.48\textwidth}
        \centering
        \scriptsize
        \caption*{non-clustered data (\texttt{option2})}
        \setlength{\tabcolsep}{4pt}
        \begin{tabular}{clcc}
            \toprule
            \# & feature (gene name) & naive & proposed (SI) \\
            \midrule
            0 & RPL34  & 0.000 & 0.962 \\
            1 & JUNB   & 0.000 & 0.974 \\
            2 & S100A4 & 0.000 & 0.000 \\
            3 & MT-CO2 & 0.000 & 0.955 \\
            4 & LTB    & 0.000 & 0.000 \\
            5 & JUN    & 0.000 & 0.981 \\
            \bottomrule
        \end{tabular}
    \end{minipage}
    \hspace*{-0.7cm}
    \label{table:pbmc3k-notdif-compare-app}
\end{table}

\begin{figure}[H]
    \centering
    \begin{subfigure}{0.48\textwidth}
        \centering
        \includegraphics[width=0.95\linewidth]{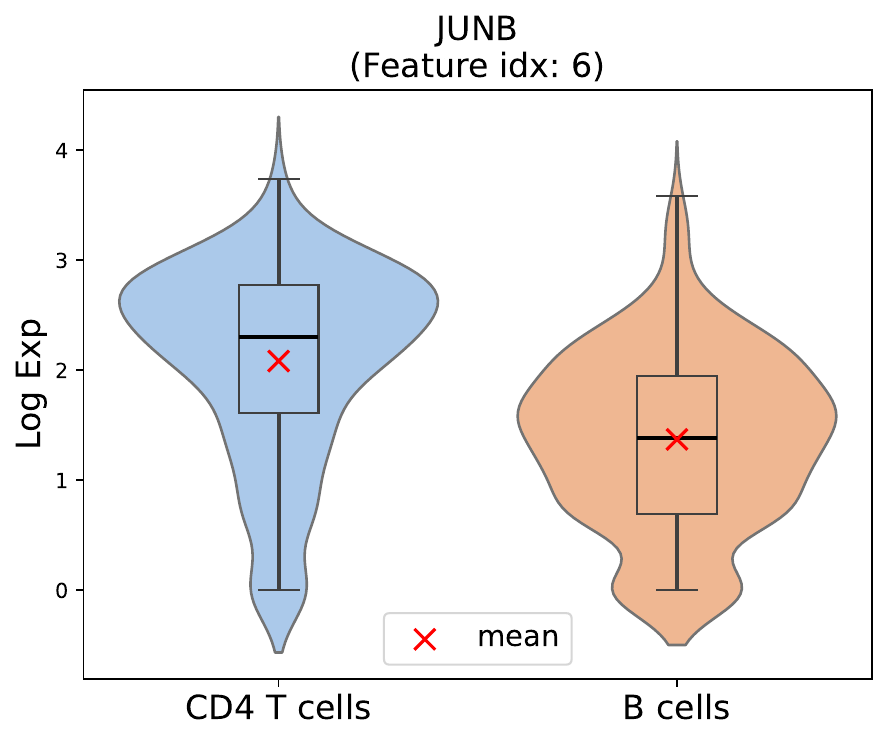}
        \caption{Distribution before clustering}
    \end{subfigure}
    \hfill
    \begin{subfigure}{0.48\textwidth}
        \centering
        \includegraphics[width=0.8\linewidth]{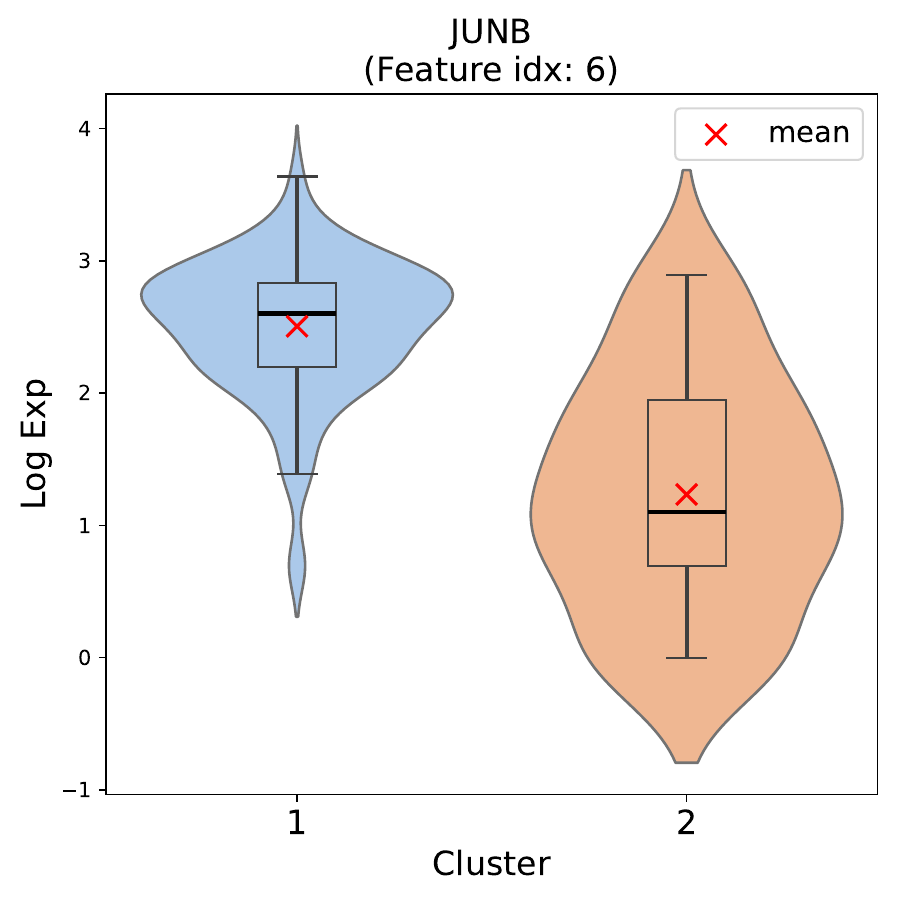}
        \caption{Distribution after clustering}
    \end{subfigure}
    \caption{
        Data distributions before and after clustering for feature \#6 (JUNB) in the clustered data.
        Cluster 1: 138 samples, Cluster 2: 55 samples.
        Although the cluster sizes are somewhat imbalanced, a clear difference in means is observed between the two clusters.
    }
    \label{fig:violin_junb}
\end{figure}

To further investigate the results, we visualize the data distributions using violin plots and box plots, focusing on representative features from the \texttt{option2} pipeline with $k$-means clustering.
We first examine feature \#6 (JUNB) in the clustered data.
Figure~\ref{fig:violin_junb} shows the data distribution before and after clustering, and Table~\ref{table:clusterd-6-pvalue} summarizes the $p$-values for each method.
\begin{table}[H]
    \centering
    \caption{$p$-values for feature \#6 (JUNB) in the clustered data}
    \label{table:clusterd-6-pvalue}
    \begin{tabular}{clccc}
        \toprule
        \# & feature (gene name) & naive & w/o-pp & proposed (SI) \\
        \midrule
        6 & JUNB & 0.000 & 0.478 & 0.000 \\
        \bottomrule
    \end{tabular}
\end{table}

From Figure~\ref{fig:violin_junb} and Table~\ref{table:clusterd-6-pvalue}, we observe the following:
\begin{itemize}
    \item The clustering successfully captures the distributional difference in JUNB expression visible in the original data.
    \item \texttt{naive} fails to control the Type I error rate and therefore cannot serve as valid evidence for detecting a cluster structure.
    \item \texttt{w/o-pp} yields a false negative due to over-conditioning, whereas the proposed method correctly detects the significant difference between clusters.
    \item This result is consistent with the established role of JUNB as a known marker gene that distinguishes CD4 T cells from B cells~\citep{koizumi2018junb,hasan2017junb}.
\end{itemize}

\begin{figure}[H]
    \centering
    \begin{subfigure}{0.48\textwidth}
        \centering
        \includegraphics[width=0.9\linewidth]{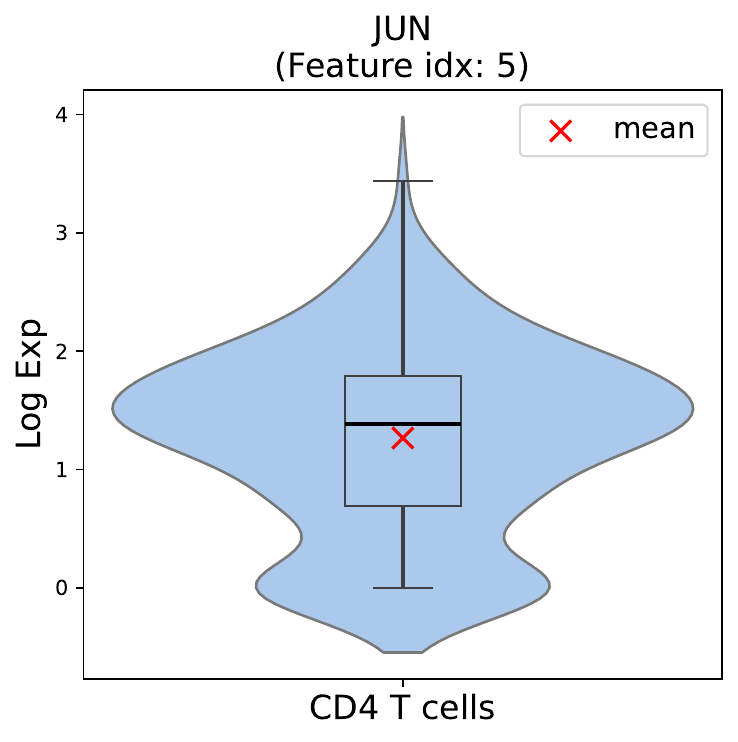}
        \caption{Distribution before clustering}
    \end{subfigure}
    \hfill
    \begin{subfigure}{0.48\textwidth}
        \centering
        \includegraphics[width=0.90\linewidth]{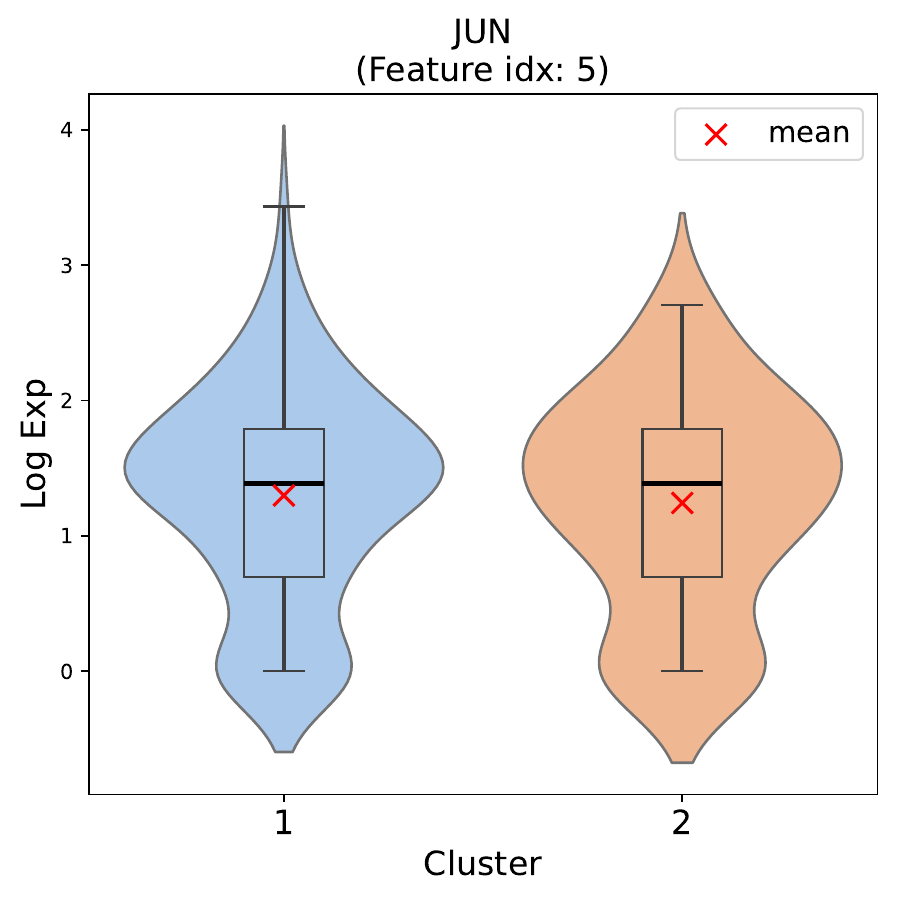}
        \caption{Distribution after clustering}
    \end{subfigure}
    \caption{
        Data distributions before and after clustering for feature \#5 (JUN) in the non-clustered data.
        Cluster 1: 123 samples, Cluster 2: 70 samples.
        Although the cluster sizes are somewhat imbalanced, no meaningful difference in means is observed between the two clusters.
    }
    \label{fig:violin_jun}
\end{figure}

We next examine feature \#5 (JUN) in the non-clustered data.
Figure~\ref{fig:violin_jun} shows the data distribution before and after clustering, and Table~\ref{table:non-clustered-5-pvalue} summarizes the $p$-values for each method.
\begin{table}[H]
    \centering
    \caption{$p$-values for feature \#5 (JUN) in the non-clustered data}
    \label{table:non-clustered-5-pvalue}
    \begin{tabular}{clccc}
        \toprule
        \# & feature (gene name) & naive & w/o-pp & proposed (SI) \\
        \midrule
        5 & JUN & 0.000 & 0.977 & 0.981 \\
        \bottomrule
    \end{tabular}
\end{table}

From Figure~\ref{fig:violin_jun} and Table~\ref{table:non-clustered-5-pvalue}, we observe the following:
\begin{itemize}
    \item The per-cluster distributions after clustering closely reflect the overall data distribution, consistent with the absence of a true cluster structure.
    \item \texttt{naive} yields a false positive, whereas both \texttt{w/o-pp} and the proposed method correctly detect the absence of a significant difference between clusters.
    \item This result is consistent with the established role of JUN as a marker gene of CD4 T cells~\citep{rincon2009ap}: since the non-clustered dataset consists entirely of CD4 T cells, no significant inter-cluster difference in JUN expression is expected.
\end{itemize}

\subsubsection{Marketing Campaign Data}
\label{app:marketing_campaign}
\begin{figure}[H]
    \centering
    \begin{subfigure}{0.48\textwidth}
        \centering
        \includegraphics[width=0.85\linewidth]{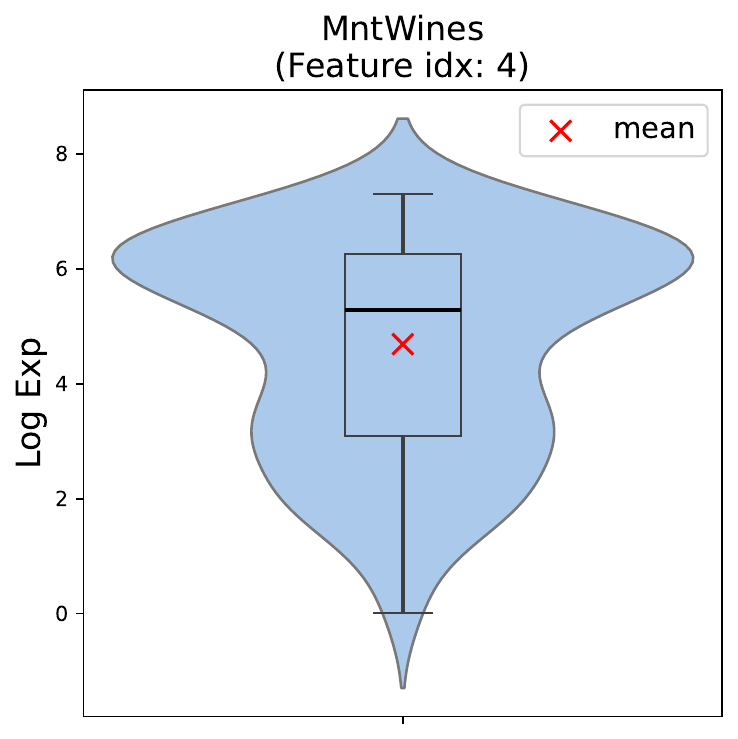}
        \vspace{0.3cm}
        \caption{Distribution before clustering}
    \end{subfigure}
    \hfill
    \begin{subfigure}{0.48\textwidth}
        \centering
        \includegraphics[width=0.9\linewidth]{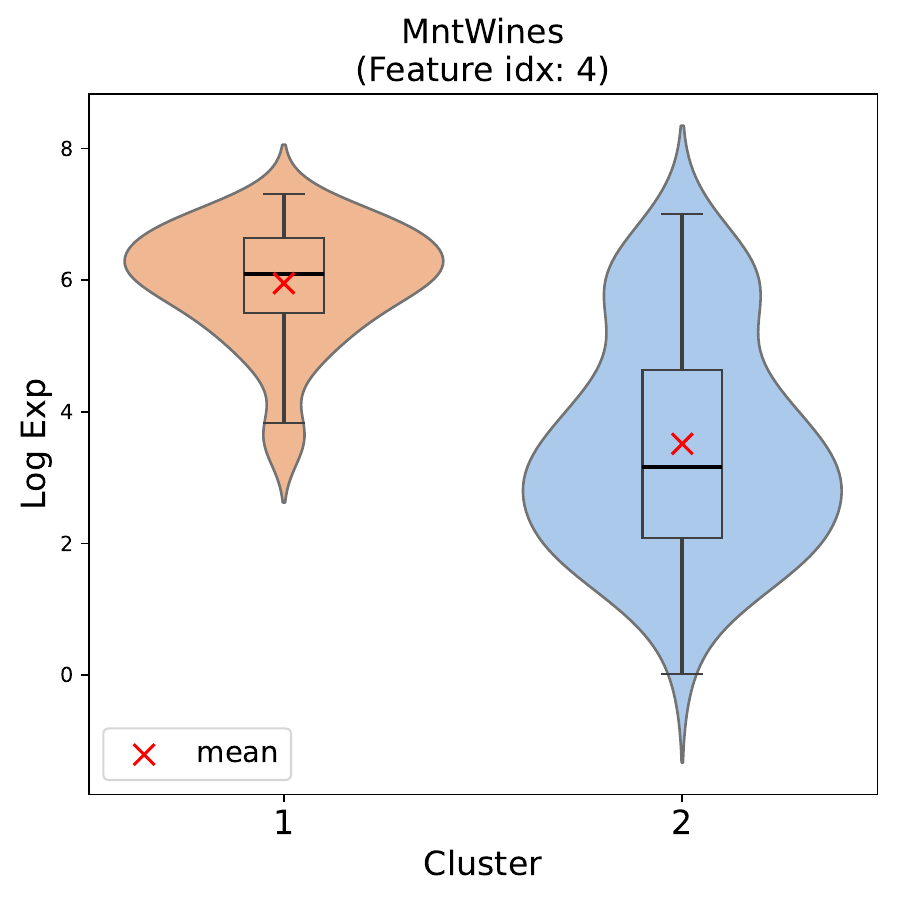}
        \caption{Distribution after clustering}
    \end{subfigure}
    \caption{
        Data distributions before and after clustering for feature \#4 (MntWines) in the clustered data.
        Cluster 1: 102 samples, Cluster 2: 90 samples.
        Two clusters of nearly equal size are formed, and a clear difference in means is observed between them.
    }
    \label{fig:violin_wine}
\end{figure}

\paragraph{Additional Results for \texttt{option1}.}
Tables~\ref{table:Marketing-dif-compare-app} shows the \texttt{option1} results for the clustered and non-clustered data, respectively (see \S\ref{sec:sec5} for the dataset and setup details).
The overall trend is consistent with the \texttt{option1} results in \S\ref{sec:sec5}: the proposed method successfully detects the cluster structure in the clustered data, and correctly suppresses false positives in the non-clustered data.
For feature \#3 (Recency) in the non-clustered data, the proposed method yields a $p$-value below the significance level, suggesting a genuine difference in purchase frequency among customers with similar incomes.
\begin{table}[H]
    \centering
    \caption{Results of \texttt{option1} for clustered and non-clustered data (Marketing Campaign).}
    \hspace*{-1.8cm}
    \label{table:Marketing-dif-compare-app}
    \begin{minipage}{0.49\textwidth}
        \centering
        \scriptsize
        \caption*{clustered data (\texttt{option1})}
        \setlength{\tabcolsep}{3pt}
        \begin{tabular}{clcc}
            \toprule
            \# & feature (spending target etc.) & naive & proposed (SI) \\
            \midrule
            4 & MntWines         & 0.000 & 0.003 \\
            5 & MntFruits        & 0.000 & 0.060 \\
            6 & MntMeatProducts  & 0.000 & 0.005 \\
            7 & MntFishProducts  & 0.000 & 0.025 \\
            8 & MntSweetProducts & 0.000 & 0.035 \\
            9 & MntGoldProds     & 0.000 & 0.000 \\
            \bottomrule
        \end{tabular}
    \end{minipage}
    \hspace{0.08\textwidth}
    \begin{minipage}{0.49\textwidth}
        \centering
        \scriptsize
        \caption*{non-clustered data (\texttt{option1})}
        \setlength{\tabcolsep}{3pt}
        \begin{tabular}{clcc}
            \toprule
            \# & feature (spending target etc.) & naive & proposed (SI) \\
            \midrule
            3 & Recency          & 0.000 & 0.000 \\
            4 & MntWines         & 0.000 & 0.070 \\
            5 & MntFruits        & 0.000 & 0.260 \\
            6 & MntMeatProducts  & 0.000 & 0.134 \\
            7 & MntFishProducts  & 0.000 & 0.292 \\
            8 & MntSweetProducts & 0.000 & 0.299 \\
            9 & MntGoldProds     & 0.000 & 0.089 \\
            \bottomrule
        \end{tabular}
    \end{minipage}
    \hspace*{-1.2cm}
    \label{table:Marketing-notdif-compare-app}
\end{table}
\clearpage
To further investigate the results, we visualize the data distributions using violin plots and box plots for representative features from the \texttt{option1} pipeline with DBSCAN clustering.
We first examine feature \#4 (MntWines) in the clustered data.
Figure~\ref{fig:violin_wine} shows the data distribution before and after clustering, and Table~\ref{table:clustered-4-pvalue} summarizes the $p$-values for each method.
\begin{table}[H]
    \centering
    \caption{$p$-values for feature \#4 (MntWines) in the clustered data}
    \label{table:clustered-4-pvalue}
    \begin{tabular}{clccc}
        \toprule
        \# & feature (spending target) & naive & w/o-pp & proposed (SI) \\
        \midrule
        4 & MntWines & 0.000 & 0.250 & 0.003 \\
        \bottomrule
    \end{tabular}
\end{table}

From Figure~\ref{fig:violin_wine} and Table~\ref{table:clustered-4-pvalue}, we observe the following:
\begin{itemize}
    \item The clustering successfully captures the distributional difference in MntWines visible in the original data.
    \item \texttt{naive} fails to control the Type I error rate and therefore cannot serve as valid evidence for detecting a cluster structure.
    \item \texttt{w/o-pp} yields a false negative due to over-conditioning, whereas the proposed method correctly detects the significant difference between clusters.
    \item Since the clustered dataset contains customers with varying income levels, the finding that there is a significant difference in wine expenditure is a naturally interpretable result.
\end{itemize}

\begin{figure}[H]
    \centering
    \begin{subfigure}{0.48\textwidth}
        \centering
        \includegraphics[width=0.9\linewidth]{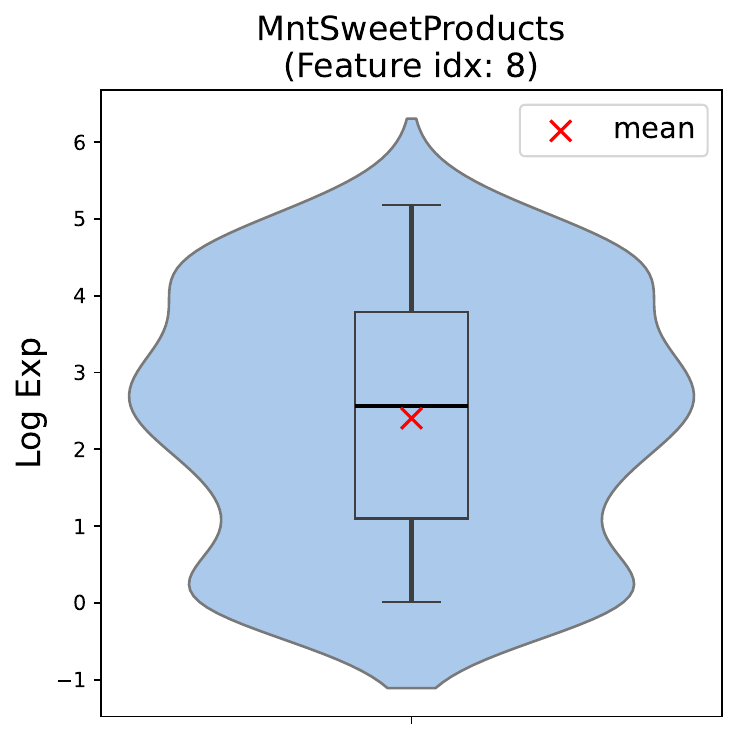}
        \caption{Distribution before clustering}
    \end{subfigure}
    \hfill
    \begin{subfigure}{0.48\textwidth}
        \centering
        \includegraphics[width=0.88\linewidth]{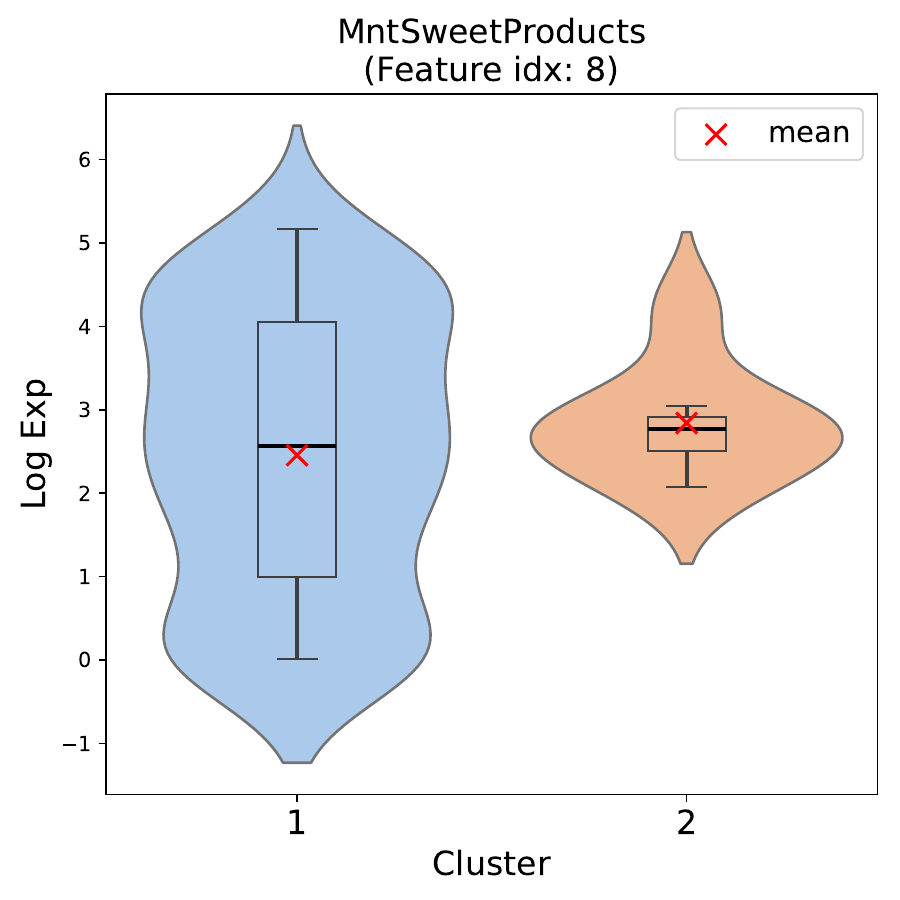}
        \caption{Distribution after clustering}
    \end{subfigure}
    \caption{
        Data distributions before and after clustering for feature \#8 (MntSweetProducts) in the non-clustered data.
        Cluster 1: 144 samples, Cluster 2: 7 samples.
        Although the cluster sizes are heavily imbalanced, no meaningful difference in means is observed between the two clusters.
    }
    \label{fig:violin_sweet}
\end{figure}

We next examine feature \#8 (MntSweetProducts) in the non-clustered data.
Figure~\ref{fig:violin_sweet} shows the data distribution before and after clustering, and Table~\ref{table:non-clustered-8-pvalue} summarizes the $p$-values for each method.
\begin{table}[H]
    \centering
    \caption{$p$-values for feature \#8 (MntSweetProducts) in the non-clustered data}
    \label{table:non-clustered-8-pvalue}
    \begin{tabular}{clccc}
        \toprule
        \# & feature (spending target) & naive & w/o-pp & proposed (SI) \\
        \midrule
        8 & MntSweetProducts & 0.000 & 0.162 & 0.299 \\
        \bottomrule
    \end{tabular}
\end{table}

From Figure~\ref{fig:violin_sweet} and Table~\ref{table:non-clustered-8-pvalue}, we observe the following:
\begin{itemize}
    \item Since the non-clustered data has no true cluster structure, DBSCAN tends to produce a highly imbalanced partition, with one cluster being much larger than the other.
    \item Both \texttt{naive} and \texttt{w/o-pp} yield false positives, whereas the proposed method correctly detects the absence of a significant difference between clusters.
    \item Since the non-clustered dataset consists of customers with similar income levels, the finding that there is no significant difference in sweet product expenditure is a naturally interpretable result.
\end{itemize}

\newpage
\section{Robustness of Type I Error Rate Control}
\label{app:robustness}
We conducted two robustness experiments using the \texttt{option2} pipeline to evaluate the Type I error rate control of the proposed method: one for the case where the variance is estimated from the same data, and one for the case where the noise follows non-Gaussian distributions.

\subsection{Estimated Variance}
\label{app:estimated_variance}
\begin{figure}[htbp]
    \centering
    \begin{subfigure}{0.46\textwidth}
        \centering
        \includegraphics[width=\textwidth]{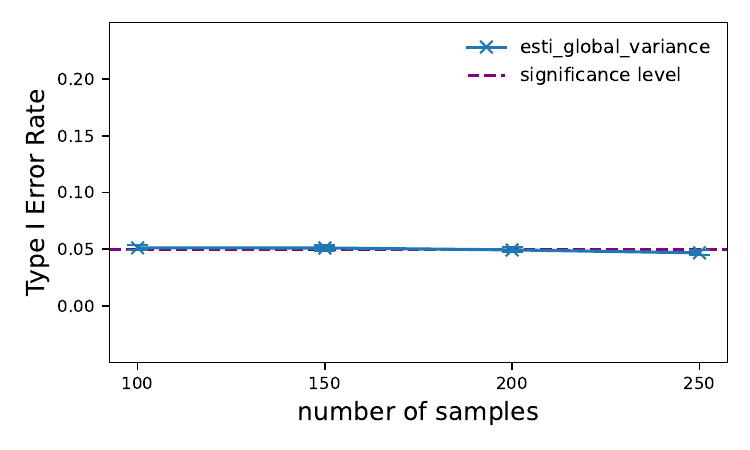}
        \subcaption{$n$ varying}
    \end{subfigure}
    \hfill
    \begin{subfigure}{0.46\textwidth}
        \centering
        \includegraphics[width=\textwidth]{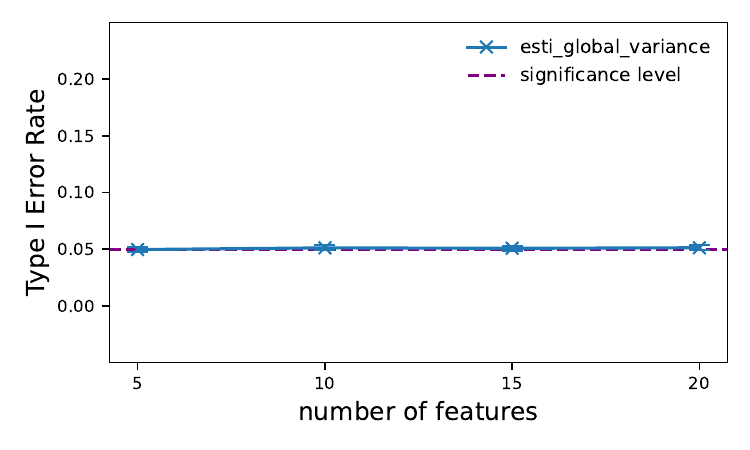}
        \subcaption{$d$ varying}
    \end{subfigure}
    \caption{
        Robustness of Type I error rate control under estimated variance.
        The proposed method successfully controls the Type I error rate below the significance level $\alpha$ even when the variance is estimated from the same data.
    }
    \label{fig:estimated_variance}
\end{figure}

In the estimated variance experiment, we used the \texttt{option2} pipeline, estimated the variance from the given data, and evaluated the Type I error rate control of the proposed method.
We considered the same two settings as in the main Type I error rate experiments (\S\ref{sec:sec5}): varying the number of samples $n$ and varying the number of features $d$.
For each setting, we generated 10,000 null datasets as described in \S\ref{sec:sec5},
and estimated the variance $\hat{\sigma}^2$ as
\begin{equation*}
    \hat{\sigma}^2 = \frac{1}{n-1}\sum_{i=1}^{n}(X_{ij^*} - \bar{X}_{j^*})^2,
\end{equation*}
where $j^*$ is the index of the selected feature and $\bar{X}_{j^*} = \frac{1}{n}\sum_{i=1}^{n} X_{ij^*}$.
We considered three significance levels $\alpha = 0.05$.
The results are shown in Figure~\ref{fig:estimated_variance}.
As can be seen, the proposed method successfully controls the Type I error rate below the significance level $\alpha$ even when the variance is estimated from the same data.

\subsection{Non-Gaussian Noise}
\begin{figure}[H]
    \centering
    \includegraphics[width=0.49\textwidth]{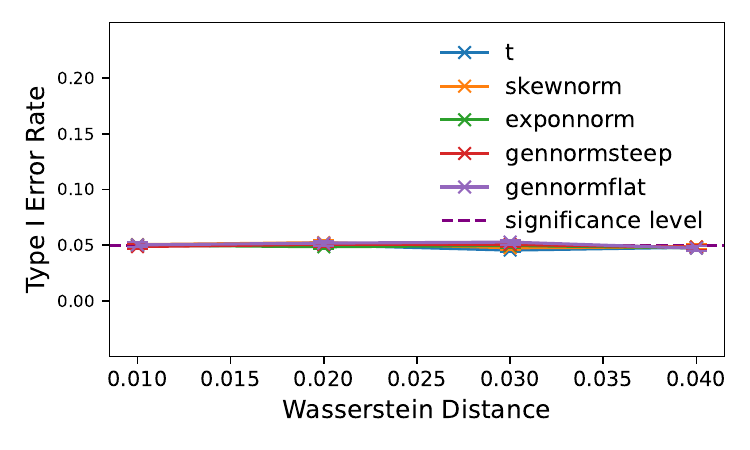}
    \caption{
        Robustness of Type I error rate control under non-Gaussian noise.
        The proposed method successfully controls the Type I error rate at the significance level $\alpha$ regardless of the Wasserstein distance from the Gaussian distribution.
    }
    \label{fig:non_gaussian}
\end{figure}

\
In the non-Gaussian noise experiment, we set $n = 100$ and $d = 10$.
We considered the following five non-Gaussian distribution families:
\begin{itemize}
    \item \texttt{skewnorm}: Skew normal distribution family.
    \item \texttt{exponnorm}: Exponentially modified normal distribution family, defined as the convolution of a normal distribution and an exponential distribution.
    \item \texttt{gennormsteep}: Generalized normal distribution family with shape parameter $\beta < 2$, which is steeper than the normal distribution.
    \item \texttt{gennormflat}: Generalized normal distribution family with shape parameter $\beta > 2$, which is flatter than the normal distribution.
    \item \texttt{t}: Student's $t$ distribution family.
\end{itemize}
Note that all of these distribution families include the Gaussian distribution as a special case, and all distributions are standardized in the experiment.
For each distribution family, we obtained distributions whose 1-Wasserstein distance from $\mathcal{N}(0,1)$ equals $l$, for $l \in \{0.01, 0.02, 0.03, 0.04\}$, and generated 10,000 null datasets accordingly.
We considered two significance levels $\alpha = 0.05$.
The results are shown in Figure~\ref{fig:non_gaussian}.
As can be seen, the proposed method successfully controls the Type I error rate at the significance level $\alpha$ even as the Wasserstein distance varies across all distribution families.

\end{document}